\documentclass[10pt,twocolumn,letterpaper]{article}

\usepackage{iccv}      
\usepackage{colortbl}

\definecolor{mygreen}{RGB}{0,103,0} 

\definecolor{cvprblue}{rgb}{0.21,0.49,0.74}
\usepackage[pagebackref,breaklinks,colorlinks,allcolors=cvprblue]{hyperref}
\usepackage{subcaption}
\def\paperID{4749} 
\def\confName{ICCV}
\def\confYear{2025}

\title{
\raisebox{-0.38ex}
{\includegraphics[width=0.56cm]{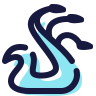}}ViT-Split: Unleashing the Power of Vision Foundation Models via \\ Efficient Splitting Heads}
\author{Yifan Li\textsuperscript{1}\thanks{Work done during interns at Bosch}, Xin Li\textsuperscript{2}, Tianqin Li\textsuperscript{2,3}, Wenbin He\textsuperscript{2}, Yu Kong\textsuperscript{1}, Liu Ren\textsuperscript{2}\\
{\normalsize \textsuperscript{1}Michigan State University}\\ 
{\normalsize \textsuperscript{2}Bosch Research North America \& Bosch Center for Artificial Intelligence (BCAI)} \\ 
{\normalsize \textsuperscript{3}Carnegie Mellon University} \\ 
{\tt\small \{liyifa11, yukong\}@msu.edu, \{xin.li9, tianqin.li2, Wenbin.He2, liu.ren\}@us.bosch.com}
}

\begin{document}
\maketitle


\begin{abstract}
Vision foundation models (VFMs) have demonstrated remarkable performance across a wide range of downstream tasks. While several VFM adapters have shown promising results by leveraging the prior knowledge of VFMs, we identify two inefficiencies in these approaches. First, the interaction between convolutional neural network (CNN) and VFM backbone triggers early layer gradient backpropagation. Second, existing methods require tuning all components, adding complexity. Besides, these adapters alter VFM features, underutilizing the prior knowledge. To tackle these challenges, we propose a new approach called ViT-Split, based on a key observation: \textbf{the layers of several VFMs, like DINOv2, can be divided into two components: an extractor for learning low-level features and an adapter for learning task-specific features}. Leveraging this insight, we eliminate the CNN branch and introduce two heads, task head and prior head, to the frozen VFM. The task head is designed to learn task-specific features, mitigating the early gradient propagation issue. The prior head is used to leverage the multi-scale prior features from the frozen VFM, reducing tuning parameters and overfitting. Extensive experiments on various tasks (e.g., segmentation, detection, depth estimation, and visual question answering) validate the effectiveness and efficiency of ViT-Split. Specifically, ViT-Split reduces training time up to $4\times$ while achieving comparable or even better results on ADE20K, compared to other VFM adapters. Codes are available: \url{https://jackyfl.github.io/vitsplit.github.io/}.
\end{abstract}
    
\vspace{-15pt}
\section{Introduction}
\label{sec:intro}

Recent studies reveal that the foundation models have the remarkable ability to acquire \textit{prior knowledge} from large-scale datasets \cite{zhou2023comprehensive}, which enhances the performance in downstream tasks. For vision tasks, vision foundation models (VFMs) acquire prior knowledge from large-scale datasets through self-supervised learning \cite{gui2024survey}, utilizing techniques such as masked image modeling (MIM) \cite{he2022masked,zhouimage,baobeit}, contrastive learning \cite{chen2020simple,he2020momentum,caron2021emerging,grill2020bootstrap}, or hybrid approaches (MIM + contrastive) \cite{oquabdinov2,assran2023self}. They also leverage vision-language alignment \cite{radford2021learning, fang2023eva} and dense prediction tasks \cite{kirillov2023segment, yang2024depth}, among others. VFMs exhibit remarkable zero-shot and transfer learning capabilities across a variety of downstream tasks, \eg, classification, detection, segmentation, monocular depth estimation (MDE), and visual question answering (VQA), \textit{etc}.

\begin{figure}[t]
    \centering
    \begin{subfigure}[b]{0.23\textwidth}
        \centering
        \includegraphics[width=1\textwidth]{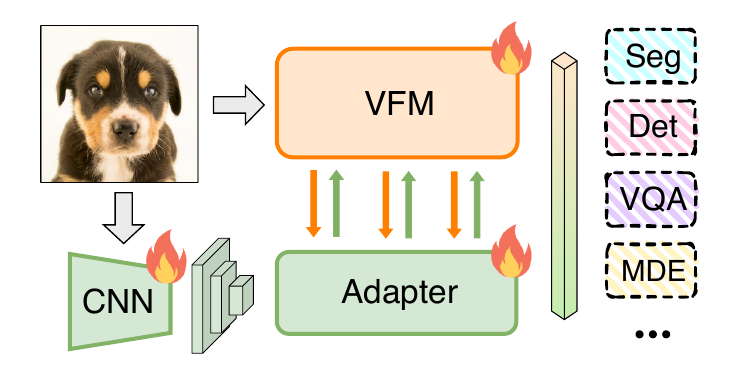}
        \caption{Previous VFM adapters.}\label{fig:adapter}
    \end{subfigure}\hfill
    \begin{subfigure}[b]{0.23\textwidth}
        \centering
        \includegraphics[width=1\textwidth]{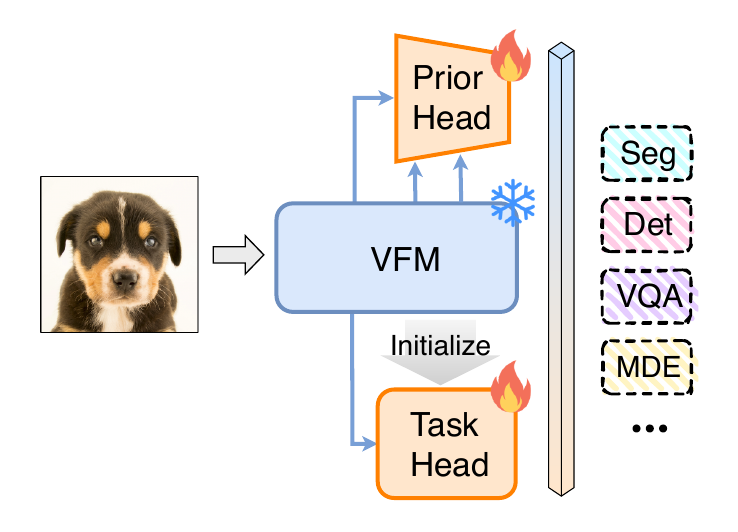}
        \caption{Ours (ViT-Split).}\label{fig:split_motivation}
    \end{subfigure}\vspace{-5pt}
    \caption{Comparison between previous VFM adapters and ours. {Previous VFM adapters integrate low-level features learned by a CNN branch into a learnable VFM through an adapter. Our method exploits VFM prior knowledge with two heads: a prior head for multi-scale prior feature learning from a frozen VFM, and a task head for task-specific feature learning, initialized by the last few layers of the VFM.}}
    \label{fig:motivation}
\end{figure}

To leverage prior knowledge from VFMs, previous VFM adapters such as ViT-Adapter \cite{chenvision} or ViT-CoMer \cite{xia2024vit} primarily adopt a two-branch architecture (see \cref{fig:adapter}). Such a design enables the adapter to integrate low-level features from a convolutional neural network (CNN) with global features from a vision transformer (ViT)-based VFM. While this architecture has demonstrated promising results across various downstream tasks, certain design aspects may affect training efficiency. From \cref{fig:adapter}, we identify two main issues of inefficiency. {First, the interaction between the CNN and ViT branches across multiple stages requires gradients to be back-propagated through all layers of the model during training. This results in increased computational and memory costs as the size of the VFM grows. Second, all components need to be tuned during training to achieve optimal performance. Specifically, for tasks like segmentation, a large head such as Mask2Former \cite{cheng2022masked} is  tuned, and its size is nearly equivalent to that of the VFM backbone.}

To address the training inefficiency issue, parameter-efficient fine-tuning (PEFT) methods are proposed to reduce training parameters. These methods include prompt-tuning approaches like VPT \cite{jia2022visual}, adapter-based methods like AdaptFormer \cite{chen2022adaptformer}, and low-rank weight tuning like LoRA \cite{hulora} or FacT \cite{jie2023fact}. However, these methods still encounter the issue of early-layer gradient back-propagation, as learnable parameters are appended to each layer’s visual tokens (prompt tuning), or low-rank weights are inserted into the layers (adapter-based methods) or added to the original weights (low-rank weight tuning). Moreover, these PEFT methods do not incorporate low-level features as VFM adapters do, and their performance is either slightly inferior to or generally on par with traditional fine-tuning. Furthermore, despite their proven effectiveness across various tasks \cite{oquabdinov2}, the pretrained prior features are not fully leveraged by either PEFT methods or the VFM adapters.

\begin{figure}
    \centering
    \includegraphics[width=1\linewidth]{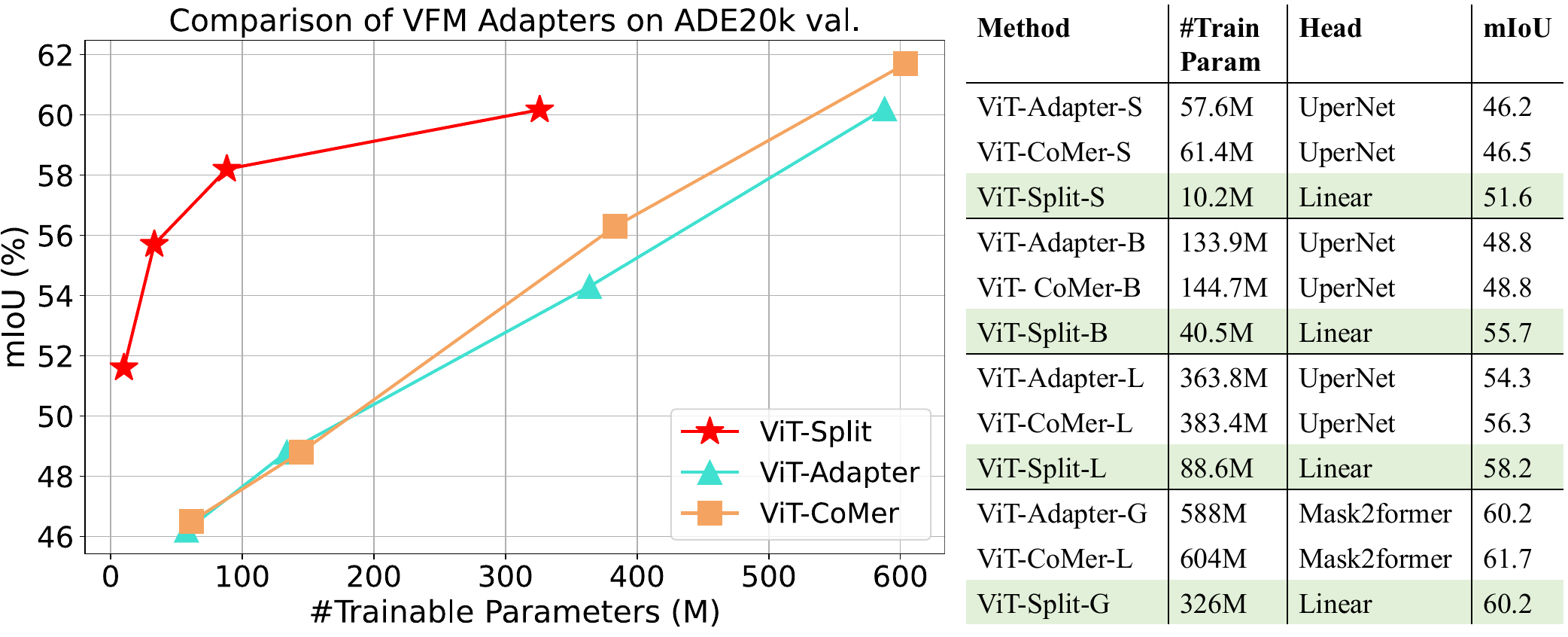}
    \caption{Comparison with previous VFM adapters (ViT-Adapter \cite{chenvision} and ViT-CoMer \cite{xia2024vit}) on ADE20K val. The results indicate that by leveraging the potential of VFMs (DINOv2 in this task), ViT-Split can achieve competitive results compared to previous VFM adapters. Notably, ViT-Split accomplishes this with only a single linear head and a small number of trainable parameters. 
    }
    \label{fig:comparison_motivation}
\end{figure}

To tackle the aforementioned challenges, we propose a method called \textit{\textbf{ViT-Split}} (see \cref{fig:split_motivation}). ViT-Split is built upon the observation that \textit{the layers of a VFM like DINOv2 \cite{oquabdinov2} can be divided into two components: a low-level feature extractor and a task-specific feature adapter}. Consequently, an additional CNN branch for local feature extraction becomes unnecessary, allowing us to remove it to resolve the early layer gradient propagation issue. Additionally, we propose a task-specific adapter, named ``task head'', tailored for downstream tasks. This adapter is initialized from the last few layers of the VFM, further avoiding gradient propagation problems in early layers. To effectively leverage prior features learned by VFM from large-scale datasets, we introduce an additional ``prior head'' that integrates multi-scale prior features instead of tuning the entire VFM. Such a head reduces the number of trainable parameters and helps mitigate overfitting in the task head (see Appendix). Additionally, we explore two layer selection strategies to identify the most relevant layer features. Experiments on segmentation task (see \cref{fig:comparison_motivation}) demonstrate that our ViT-Split, using only a single linear head, can achieve competitive performance compared with previous VFM adapters with larger segmentation heads like Mask2former \cite{cheng2022masked} or UperNet \cite{xiao2018unified}, while tuning fewer parameters and reducing training time (see \cref{fig:time_comparison}). 

Furthermore, ViT-Split is both \emph{adaptive and memory efficient for multiple tasks} (see \cref{fig:inference_comparison}). Previous VFM adapters require separate modules (VFM+CNN+adapter+heads) for each task, leading to high computational and memory overhead. In contrast, ViT-Split shares a pre-trained VFM backbone, requiring only a task-specific adapter and the corresponding task head to be learned. Our approach introduces a new paradigm for designing both computation and memory efficient VFM adapters across multiple tasks. In summary, the contributions of this paper are threefold:
\begin{itemize}
    \item We observe that several VFMs, especially DINOv2, can be divided into two distinct components:  an extractor for learning low-level features and an adapter for learning task-specific features.
    \item We propose an efficient and effective adapter ViT-Split for VFMs. Specifically, ViT-Split introduces two heads, a task head and a prior head. The task head is for learning task-specific features. The prior head is a lightweight CNN for extracting multi-scale prior features from a frozen VFM. We also explore two layer selection methods for selecting prior features from all the layers: uniform sampling and sparse gate.
    \item We perform extensive experiments and detailed ablations on various downstream tasks to validate the efficiency and effectiveness of our method, including segmentation, detection, MDE, and VQA.
\end{itemize}





\section{Related Work}
\subsection{Vision foundation models}
Vision foundation models (VFMs) \cite{awais2023foundational} are trained on large-scale datasets in a self-supervised, weakly-supervised, or supervised manner, making them adaptable to a wide range of downstream tasks. Benefiting from the scalability of the transformer architecture, recent ViT-based \cite{dosovitskiy2020image} VFMs demonstrate remarkable zero-shot and transfer ability across various downstream tasks. Self-supervised pretraining paradigm learns discriminative features solely from vision data at the image and pixel level, including contrastive learning (MoCo \cite{he2020momentum}, SimCLR \cite{chen2020simple}), masked image modeling (BEiT \cite{baobeit}, MAE \cite{he2022masked}, iBoT \cite{zhouimage}) or hybrid approaches (DINOv2 \cite{oquabdinov2}, I-JEPA \cite{assran2023self}). Weakly-supervised pretraining paradigm leverages text guidance, aligning visual representations with language space, such as CLIP \cite{radford2021learning}, ALIGN \cite{jia2021scaling}, EVA2 \cite{fang2023eva}, SigLip \cite{zhai2023sigmoid}, \etc. Supervised pretraining paradigm learns from different task labels, such as classification (DeiT \cite{touvron2021training}), segmentation (SAM \cite{kirillov2023segment}), and monocular depth estimation (DAM \cite{yang2024depth}), \etc.

\subsection{PEFT and VFM adapters}

As the size of transformer-based foundation models continues to grow, such as large language models in  language ~\cite{brown2020language,zhao2023survey}, large vision models in vision \cite{dehghani2023scaling,yucoca2022}, and multi-modal large language model \cite{chen2024internvl,bai2023qwen} for multi-modal learning,  training efficiency becomes increasingly crucial. To address this challenge, PEFT methods have gained significant popularity in recent years.

Current PEFT approaches for vision \cite{yu2024visual} generally fall into three categories: prompt tuning, adapter tuning, and parameter tuning. Prompt tuning involves learning a small number of prompt tokens, either in the first layer (CoOp ~\cite{zhou2022learning}, CoCoOp \cite{zhou2022conditional}) or in every layer (VPT \cite{jia2022visual}), making it lightweight and easy to implement. Adapter tuning inserts additional blocks into a frozen model either in a sequential manner (Res-adapt \cite{rebuffi2017learning}, ST-Adapter \cite{pan2022st}) or in parallel (AdaptFormer \cite{chen2022adaptformer}, ConvPass \cite{jie2022convolutional}, LoSA \cite{mercea2024time}), which shows good adaptability and generalizability. Parameter tuning modifies part of the model parameters, either by adjusting the weight (LoRA \cite{hulora}, FacT \cite{jie2023fact}) or tuning the bias (Bitfit \cite{zaken2022bitfit}), resulting in effective and efficient tuning.

Current VFM adapters (ViT-Adapter \cite{chenvision}, ViT-CoMer \cite{xia2024vit}) aim to enhance full fine-tuning performance by incorporating the inductive bias from the CNN branch with spatial prior. These adapters typically require tuning the whole backbone to achieve optimal performance, resulting in better performance than PEFT methods. The interaction between CNN and ViT features is achieved through cross-attention \cite{chenvision}, self-attention \cite{xia2024vit} or mixed \cite{zhang2025memory} across several layers. By contrast, our ViT-Split keeps the entire backbone frozen, introducing two lightweight heads for separate tuning, which is efficient and effective across various tasks.

\section{Method}

\subsection{The observation in VFMs}\label{observation}

\begin{figure}
    \centering
    \includegraphics[width=1\linewidth]{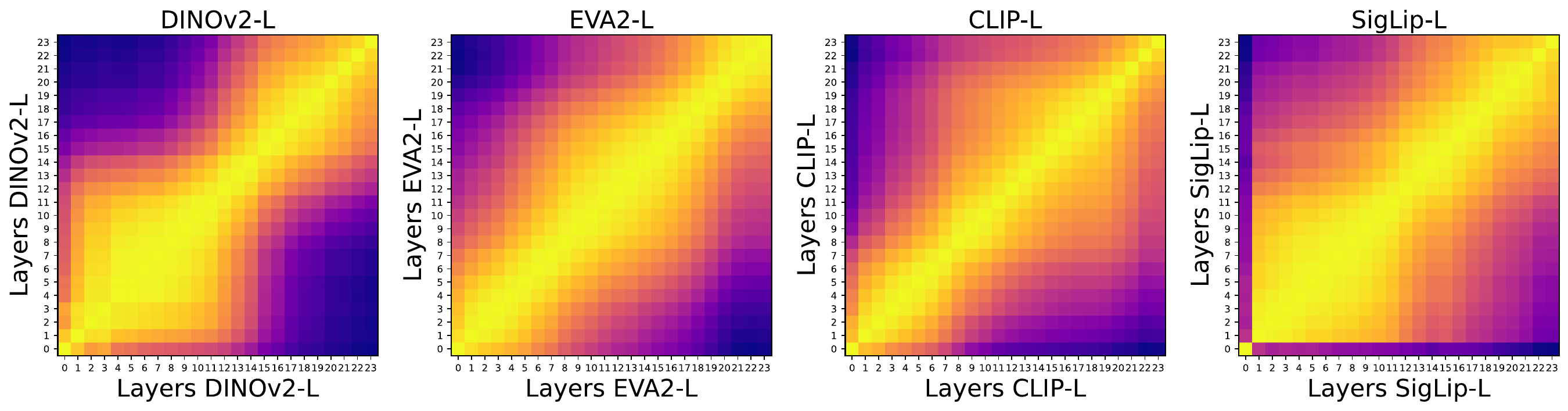}
    \caption{The CKA comparison of layer features across different VFMs, including a self-supervised method DINOv2-L \cite{oquabdinov2}, and three image-text alignment methods EVA2-L \cite{fang2023eva}, CLIP-L \cite{radford2021learning} and SigLip-L \cite{zhai2023sigmoid}. For most of these VFMs, especially DINOv2, the features in the early and later layers show distinct similarities within their respective groups. }
    \label{fig:cka_comparison}
\end{figure}

We observe that in some VFMs, the layers can be broadly partitioned into two groups with similar features: the earlier and later layers. First, we plot the Centered Kernel Alignment (CKA) \cite{kornblith2019similarity} across different layers for several VFMs, as shown in \cref{fig:cka_comparison}. The results reveal that features in the earlier layers are more similar to each other, as are those in the later layers, particularly in DINOv2 \cite{oquabdinov2}. We attribute this phenomenon to the ``encoder-decoder'' architecture intrinsic to VFMs: the earlier layers function as an encoder (feature extractor) to capture features from the visual data, while the later layers act as a decoder (task-specific adapter) that generates features for downstream tasks.

\begin{figure}
    \centering
    \includegraphics[width=1\linewidth]{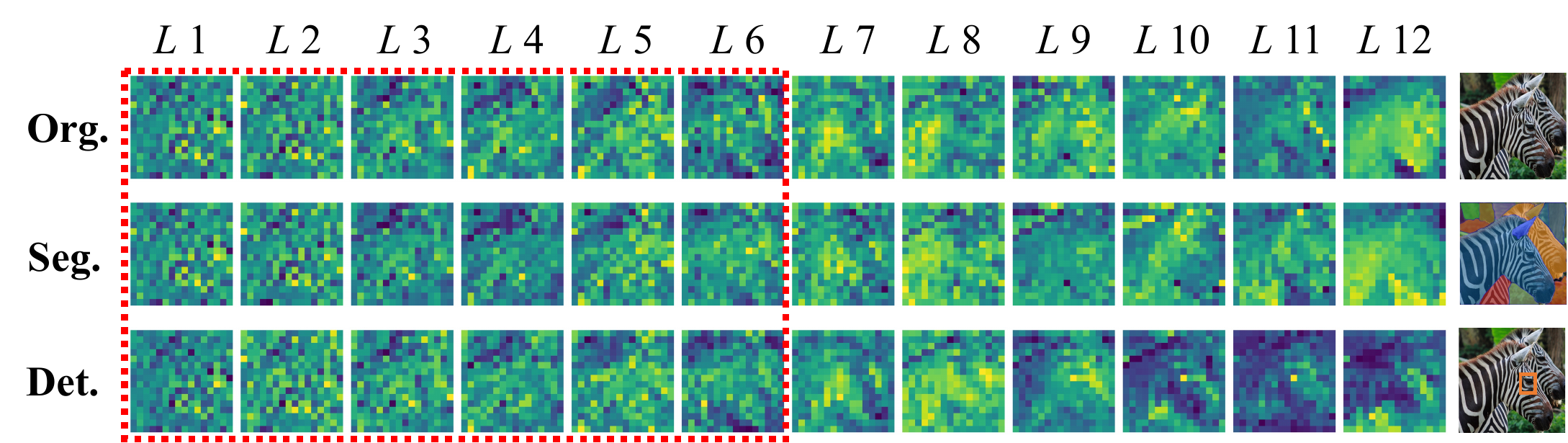}
    \caption{Comparison of DINOv2-S layer features  across different tasks, including pretraining (org.), segmentation (seg.), and detection (det.). Notably, the segmentation and detection models are fine-tuned from the DINOv2-S. The features within the red dotted boxes across the three tasks exhibit similar patterns, emphasizing detailed representations. In the later layers, however, the features diverge, becoming more specialized for each task.}
    \label{fig:layer_feature}
\end{figure}

A research question is raised: \textit{what do these two groups of layers actually learn}? To answer this question, we visualize the features of each layer in DINOv2-S (\cref{fig:layer_feature}) using the first channel of the visual tokens. To further explore feature differences across downstream tasks, we fine-tune the same DINOv2-S on segmentation and detection tasks by adding a linear head and a Mask R-CNN \cite{he2017mask} head, respectively. As shown in \cref{fig:layer_feature}, we observe that in the early layers (say layer 1-6), all three models exhibit similar feature patterns, focusing more on low-level features like texture and edges. This observation is also supported in \cite{raghu2021vision}, which demonstrates that ViT can learn low-level features through large-scale pretraining. While in the later layers, the features diverge for different tasks. Specifically, for the original DINOv2 and segmentation features (row 1 and row 2), the focus shifts towards the semantic information of objects. Whereas in the detection task, the feature attention gradually moves to the object corners or edges (row 3, L7-L12). We attribute this phenomenon to the intrinsic characteristics of each task: DINOv2's pretraining objective is to reconstruct missing parts of the original features, which requires the semantic level understanding as the segmentation task does. In detection, the goal is to predict object bounding boxes, which necessitates focusing more on the corners. This phenomenon also highlights the difference between dense prediction and detection task.

Based on the findings, we divide layers of VFMs into two groups with similar features: a feature extractor for learning low-level features and a task-specific adapter for learning task-related features.

\subsection{ViT-Split} 

\begin{figure}
    \centering
    \includegraphics[width=0.95\linewidth]{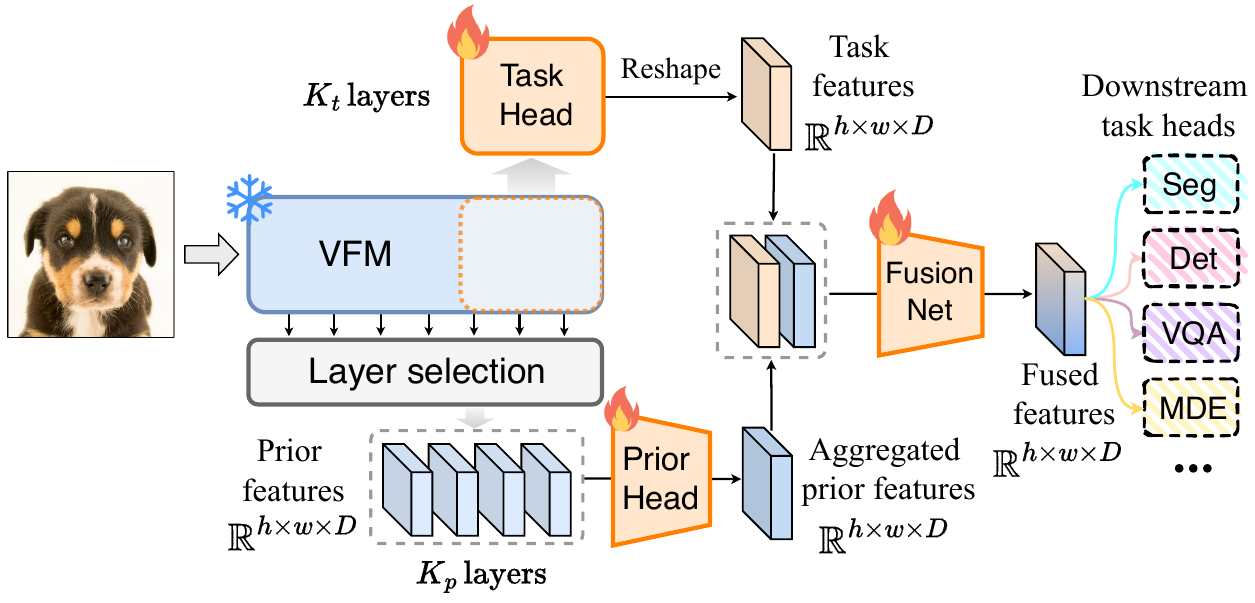}\vspace{-5pt}
    \caption{The framework of  ViT-Split. ViT-Split introduces two splitting heads, one prior head for aggregating multi-scale prior features from VFM and a task head for learning task-specific features. These features are then combined using a fusion network, enabling effective performance across various downstream tasks.} 
    \label{fig:framework}
\end{figure}

The framework of ViT-Split is illustrated in \cref{fig:framework}, which includes three trainable components: a task head, a prior head and a fusion net. The task head, initialized with the last few layers of the VFM, is designed to learn task-specific features. The prior head integrates multi-scale prior features from the VFM, which are learned from large-scale, diverse datasets. Finally, the fusion net combines both task-specific and prior features to support various downstream tasks.

When an input image with a shape of $H\times W$ is fed into a frozen VFM (\eg, DINOv2), $h\cdot w$ vision tokens with $D$ channels will be obtained from each layer. The vision tokens from ($L-K_t$) layer are passed through a task head, which is copied from the last $K_t$ layers of the VFM, where $L$ is the number of the total layers. The task features are then reshaped to $h\times w\times D$. Meanwhile, $K_p$ layers of prior features from the frozen VFM are sampled using selection strategies, then concatenated and reshaped into a feature map of size $h \times w \times (K_p \cdot D)$. The feature map is then passed through a prior head, a two-layer CNN, resulting in a prior feature map of shape $h\times w \times D$. Finally, the task and prior feature maps are concatenated along the channel dimension and fused by a fusion net, which has a similar architecture to the prior head. The final fusion feature map is provided for different downstream heads.

\textbf{Task Head.} Based on the observation in \cref{observation} that early layers of VFMs are capable of learning low-level features which are similar for different tasks, we avoid fine-tuning the entire backbone by sharing these early layers. Meanwhile, to retain the prior features of the VFM, we replicate the final $K_t$ layers separately, utilizing them as a task-specific adapter for downstream tasks. The hyperparameter $K_t$ controls the adapter's size, balancing between model capacity and training efficiency.

We observe that the benefits of increasing $K_t$ diminish, particularly for segmentation tasks, allowing us to choose a smaller $K_t$ to enhance efficiency (see hyper-parameter analysis in Appendix). Additionally, we find that a large segmentation head may be unnecessary, as the task-specific head is sufficient to capture the downstream dataset's specific knowledge. Let the features from the ($L-K_t$)-th layer of the VFM be denoted as $f_{L-K_t}$. Consequently, the task-specific features are given by:
\begin{equation}
f_{t} = g_{\theta_{t}}(f_{L-K_t}), 
\end{equation}
where $g_{\theta_t}$ represents the task head. After obtaining the task feature $f_t \in \mathbb{R}^{(h\cdot w+1)\times D}$, we drop the class token and reshape it from the sequence dimension to form a feature map $f'_t \in \mathbb{R}^{h\times w\times D}$.

\textbf{Prior Head}. The prior features learned by VFMs have demonstrated strong performance across a range of downstream tasks \cite{radford2021learning, oquabdinov2}. However, most current VFM adapters and PEFT methods modify these prior features during training. In contrast, our ViT-Split approach fully leverages the prior knowledge embedded in the multi-scale features of the VFM through a dedicated prior head. Our rationale for utilizing these prior features is to harness the knowledge learned by VFMs to enhance task-specific features while mitigating the risk of overfitting downstream tasks. 

\begin{figure}
    \centering
    \includegraphics[width=0.94\linewidth]{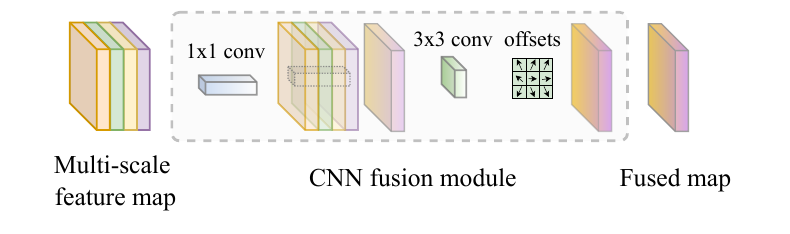}\vspace{-5pt}
    \caption{The illustration of the  CNN fusion architecture. It is used to fuse multi-scale feature maps and serves as the architecture for both the prior head and fusion net. This module consists of two CNN layers: a 1×1 convolution layer followed by a 3×3 deformable convolution layer. }
    \label{fig:fusion_cnn}
\end{figure}

Specifically, the architecture of the prior head is shown in \cref{fig:fusion_cnn}, consisting of two CNN layers, a 1$\times$1 convolution layer and a 3$\times$3 deformable convolution layer. The 1$\times$1 convolution layer is used to compress the channels of the multi-scale feature maps, providing efficiency when dealing with larger scales. Meanwhile, the deformable convolution layer \cite{dai2017deformable} enhances low-level features and models geometric transformations within the feature map.

\textbf{Layer Selection}. \emph{How to select suitable prior features from all the VFM layers?} To address this, we explore two techniques for selecting $K_p$ layers from a total of $L$ layers: uniform sampling and sparse gate. We delineate sparse gate in the Appendix.
Uniform sampling involves selecting $K_p$ prior features uniformly from $L$ layers. This design is motivated by two factors: first, mitigating the high similarity between features of neighboring layers (see \cref{fig:cka_comparison}), and second, promoting greater diversity among the selected features. Specifically, the set of sampled indices, $\mathcal{S}$, is defined as follows:
\begin{equation}\small
    \delta = \frac{L-b-1}{K_p-1}, 
    \mathcal{S} = \{b+{\rm{round}}(i\cdot\delta) | i=0, ..., K_p-1\},
\end{equation}
where $b$ is the starting index, used to skip the first few layers, as these layers tend to contain more noise. In most experiments, we set $b=2$ or $b=3$. $\rm{round}$ indicates the rounding to the nearest integer, and $\delta$ represents the sampling interval. 

After obtaining the selected prior features $f^i_p\in \mathbb{R}^{(h\cdot w + 1) \times D}, i=\{0,...,K_{p}-1\}$, we drop the class tokens, reshape and concatenate them to a multi-scale prior feature map $f_p\in \mathbb{R}^{h\times w \times (K_{p}\cdot D)}$. Finally, the aggregated prior map $f'_p\in \mathbb{R}^{h\times w \times D}$  can be denoted as:
\begin{equation}
    f'_p = g_{\theta_p}(f_p),
\end{equation}
where $g_{\theta_p}$ is the prior head.

\textbf{Fusion net}. Fusion net is utilized to fuse prior feature map $f'_p$ and the task-specific feature map $f'_t$ for different downstream tasks. This network has a similar architecture as the prior head (see \cref{fig:fusion_cnn}). Let $[f'_p; f'_t]\in \mathbb{R}^{h\times w \times (2D)}$ be the concatenated feature map of $f'_p$ and $f'_t$ along the channel dimension. The rationale of using concatenation to fuse two feature maps is to preserve more information (see \cref{tab:ablation}). The final fused map $f_o\in \mathbb{R}^{h\times w \times D}$ is given by:
\begin{equation}
    f_o = g_{\theta_f}([f'_p; f'_t]),
\end{equation}
where $g_{\theta_f}$ is the fusion net.

We then apply different transformations based on the type of downstream task. Specifically, for the segmentation task, we upsample $f_o$ by a factor of 4 using two transposed convolution layers.  For the detection task, we transform $f_o$ into four scales, \ie, $4\times$, $2\times$, $1\times$ and $0.5\times$ to match the input requirements of the detection head (MaskRCNN). For the VQA task, we reshape $f_o$ along the sequence dimension to ${(h\cdot w)\times D}$ for the LLM decoder.

\section{Experiments}

We conduct experiments on three tasks, semantic segmentation, object detection, and VQA, using well-established benchmarks, \textit{e.g.}, COCO \cite{lin2014microsoft}, ADE20K \cite{zhou2019semantic}, CityScapes \cite{cordts2016cityscapes}, among others. We also present MDE results in the Appendix.
Next, we perform ablation studies to further evaluate ViT-Split’s performance. A uniform selection strategy is applied to all experiments in this section, while results for the sparse gate are provided in the Appendix.

\begin{table}[t]
\centering 
\small
\setlength{\tabcolsep}{3.5pt} 
\begin{tabular}{l | ccccc}
\hline
{Method} &  Head    & \#Train Param     & mIoU     & Iters     \\
\hline
PVT-S~\cite{wang2022pvtv2}                 & UperNet  & 54.5M       & 43.7     & 160k    \\
Swin-T~\cite{liu2021swin}                           & UperNet  & 59.9M       & 44.5     & 160k    \\
Twins-SVT-S~\cite{chu2021twins}                     & UperNet & 54.4M       & 46.2     & 160k    \\
ViT-S~\cite{li2021benchmarking}                         & UperNet   & 53.6M       & 44.6     & 160k    \\
LoSA-S~\cite{mercea2024time}                         & UperNet   & 54.9M       & 45.8     & 160k    \\ 
ViT-Adapter-S~\cite{chenvision}                     & UperNet  & 57.6M       & 46.2     & 160k    \\
ViT-CoMer-S~\cite{xia2024vit}                  & UperNet &    61.4M       &    {46.5}   &   160k   \\
DINOv2S\textsuperscript{$\ddagger$}~\cite{oquabdinov2} & UperNet & 52.2M & {50.6} & 40k \\
DINOv2-S\textsuperscript{$\ddagger$}~\cite{oquabdinov2} & Linear & 22.1M & {49.6} & 40k \\
\rowcolor[HTML]{E2F0D9}
ViT-Split-S\textsuperscript{$\ddagger$} (ours) & Linear & 10.2M & \textbf{51.6} & 40k \\
\hline
Swin-B~\cite{liu2021swin}                          & UperNet    & 121.0M      & 48.1     & 160k    \\
Twins-SVT-L~\cite{chu2021twins}                         & UperNet  & 133.0M      & {48.8}     & 160k    \\
ViT-B~\cite{li2021benchmarking}                          & UperNet   & 127.3M      & 46.1     & 160k    \\
LoSA-B~\cite{mercea2024time}                         & UperNet   & 131.2M       & 47.3     & 160k    \\ 
ViT-Adapter-B~\cite{chenvision}                   & UperNet  & 133.9M      & {48.8}     & 160k    \\
ViT-CoMer-B~\cite{xia2024vit}                & UperNet    &      144.7M       &  {48.8}     &  160k     \\
DINOv2-B\textsuperscript{$\ddagger$}~\cite{oquabdinov2} & UperNet & 120.7M & {54.8} & 40k \\
DINOv2-B\textsuperscript{$\ddagger$}~\cite{oquabdinov2} & Linear & 91.4M & {53.8} & 40k \\
\rowcolor[HTML]{E2F0D9}
ViT-Split-B\textsuperscript{$\ddagger$} (ours) & Linear & 40.5M & \textbf{55.7} & 40k \\
\hline
Swin-L\textsuperscript{$\dagger$}~\cite{liu2021swin}                   & UperNet         & 234.0M      & 52.1     & 160k    \\
LoSA-L\textsuperscript{$\dagger$}~\cite{mercea2024time}                         & UperNet   & 338.5M       & 53.0     & 160k    \\ 
ViT-Adapter-L\textsuperscript{$\dagger$}~\cite{chenvision}             & UperNet         & 363.8M      & 53.4     & 160k    \\
ViT-CoMer-L\textsuperscript{$\dagger$}~\cite{xia2024vit}\textsuperscript              & UperNet     & 383.4M      & {54.3}     &  160k    \\
DINOv2-L\textsuperscript{$\ddagger$}~\cite{oquabdinov2}\textsuperscript              & UperNet   & 341.2M      & {57.1}     &  40k    \\
DINOv2-L\textsuperscript{$\ddagger$}~\cite{oquabdinov2}\textsuperscript              & Linear   & 312.9M      & {56.2}     &  40k    \\
\rowcolor[HTML]{E2F0D9}
ViT-Split\textsubscript-L\textsuperscript{$\ddagger$} (ours) & Linear & 88.6M & \textbf{58.2} & 40k \\
\hline
\end{tabular}
\caption{\textbf{Semantic segmentation results on the ADE20K val with 512*512 resolution image. } $\ddagger$ represents the DINOv2 initialization. ``$\dagger$'' denotes the use of ImageNet-22K pre-trained weight, while the default is to use ImageNet-1K pre-training. }
\label{tab:seg_ADE20K}
\end{table}

\begin{figure*}
    \centering
    \begin{minipage}[b]{0.6\linewidth}
        \small
        \setlength{\tabcolsep}{4pt} 
        \scalebox{0.73}{
            \begin{tabular}{l | cccccc}
            \hline
            {Method} &  Head    & \#Train     & mIoU (SS/MS)   & Pretrain     & Extra Pre-train & Iters\\
            \hline
            ConvNeXt-XL \cite{liu2022convnext} & Mask2former \cite{cheng2022masked} & 588M & 57.1/58.4 & IN-22k & COCO-Stuff & 80k \\
            Swin-L \cite{liu2021swin} & Mask2former \cite{cheng2022masked} & 434M & 57.3/58.3 & IN-22k & COCO-Stuff & 80k \\
            SwinV2-G \cite{liu2022swin} & UperNet \cite{xiao2018unified} & 3B & 59.3/59.9 & IN-22K & Ext-70M & 160k \\
            Swin-L \cite{liu2021swin} & MaskDINO \cite{li2023mask} & 223M & 59.5/60.8 & IN-22k & Object365 & 160k\\
            ViT-CoMer-L \cite{xia2024vit} & Mask2former \cite{oquabdinov2} & 604M & \textbf{61.7}/\textbf{62.1} & MM, BEiTv2 & COCO-Stuff & 80k \\
            \hline
            DINOv2-L\textsuperscript{$\ddagger$}~\cite{oquabdinov2} & Linear & 312.9M & 58.1/58.5 & LVD-142M & -- & 40k \\
            \rowcolor[HTML]{E2F0D9}
            ViT-Split-L\textsuperscript{$\ddagger$} (ours) & Linear & 86.2M & 59.0/59.6 & LVD-142M & -- & 40k \\
            \hline
            ViT-Adapter-G\textsuperscript{$\ddagger$*}~\cite{chenvision}  & Mask2former \cite{oquabdinov2} & 588M  & --/60.2 & LVD-142M & --  & 80k    \\
            \rowcolor[HTML]{E2F0D9}
            ViT-Split-G\textsuperscript{$\ddagger$}~(ours) & Linear & 326M & \underline{60.2}/\underline{60.8} & LVD-142M & -- & 50k \\
            \hline
            \end{tabular}\vspace{-10pt}
        }
        \captionof{table}{\textbf{Compared with previous SOTA segmentic segmentation methods on ADE20K val with 896*896 resolution image. }$\ddagger$~ are initialized with DINOv2. * is implemented without tuning the whole backbone \cite{oquabdinov2}. ``MS'' means multi-scale testing. ``MM'' indicates multi-modal pretraining.}
        \label{tab:seg_ADE20K_896}
    \end{minipage}
    \hfill
    \begin{minipage}[b]{0.38\linewidth}
        \includegraphics[width=1.04\linewidth]{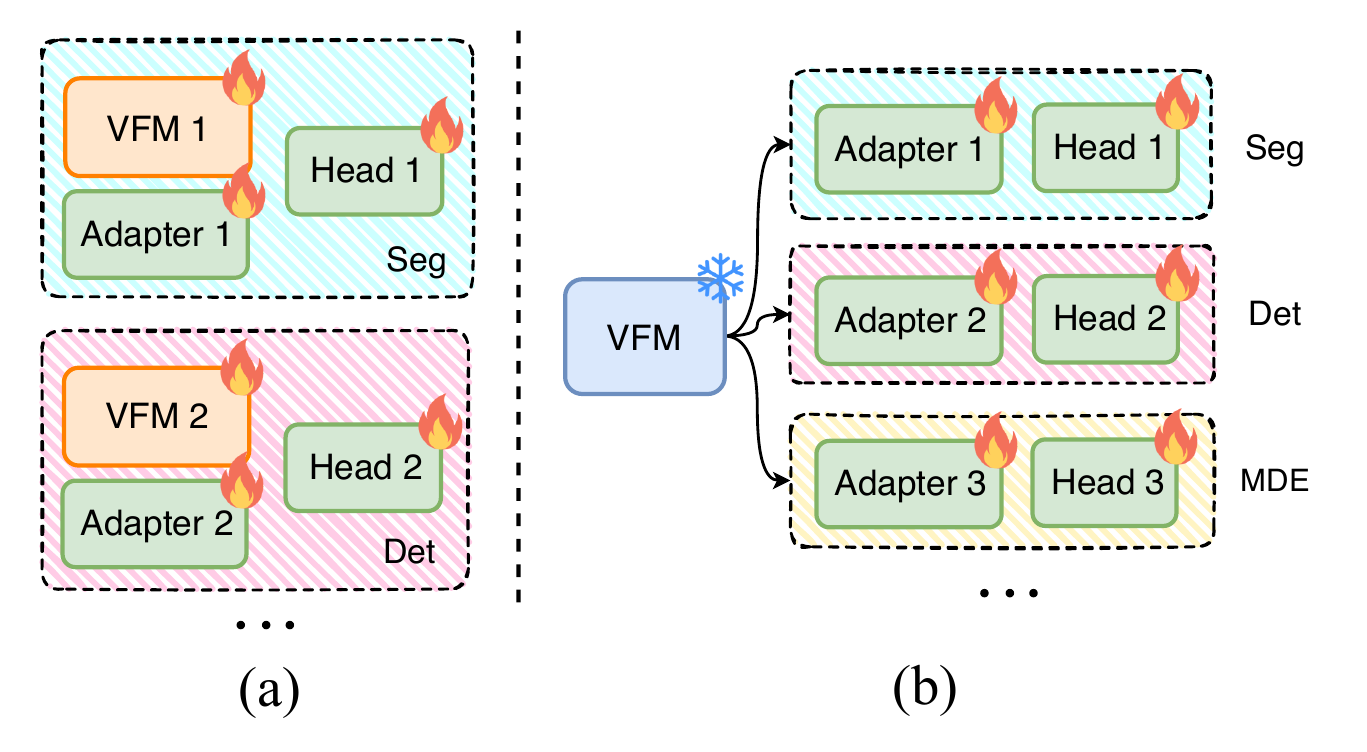}\vspace{-10pt}
        \caption{Inference comparison: (a) Previous VFM adapters vs. (b) Our ViT-Split. ViT-Split is  efficient during inference for multiple tasks.}
        \label{fig:inference_comparison}
    \end{minipage}
\end{figure*}

\subsection{Semantic segmentation}\label{sec:seg}

\textbf{Settings}. We conduct the semantic segmentation task on ADE20K \cite{zhou2019semantic} and Cityscapes \cite{cordts2016cityscapes}, using MMSegmentation \cite{contributors2020mmsegmentation}. We employ AdamW \cite{loshchilov2017decoupled} with a learning rate of 2e-4 and a weight decay of 1e-2. The training process uses a total batch size of 16. The learning rate for the task head is further reduced by a factor of 0.1. Unlike previous baselines, we use a simple linear head with two-layer deconvolutional blocks ($\times$4) for segmentation, with a total of 40k iterations (50k for DINOv2-g). We provide the hyper-parameter analysis of $K_p$ and $K_t$ in the Appendix.

\textbf{ADE20K val with 512$\times$512 image.} As shown in \cref{tab:seg_ADE20K}, we can see that our ViT-Split surpasses all other baselines on ADE20K with 512$\times$512 resolution input image by fully leveraging the potential of the VFM. The results demonstrate the superiority of the DINOv2 compared to ImageNet pretrained models. Additionally, ViT-Split requires tuning only about 1/5 to 1/4 of the parameters and trains for just 1/4 of the iterations compared to previous baselines. The parameter efficiency is because of: 1) the efficient adaptation architecture of ViT-Split and 2) the lightweight linear head. The fast convergence speed attributes to effective utilization of the prior knowledge embedded in VFMs. Moreover, compared to fine-tuning the entire DINOv2 baseline, our ViT-Split adjusts only 1/4 to 1/2 of the parameters while achieving an average improvement of 2\% across three model sizes.  Since most tunable parameters come from the tuned head, which represents a small portion of the entire VFM, the overall parameter count for tuning remains low. The performance gains can be attributed to the utilization of the multi-scale prior features from the VFM. 

\begin{figure}[t]
    \centering
    \begin{subfigure}[b]{0.23\textwidth}
        \centering
        \includegraphics[width=1\textwidth]{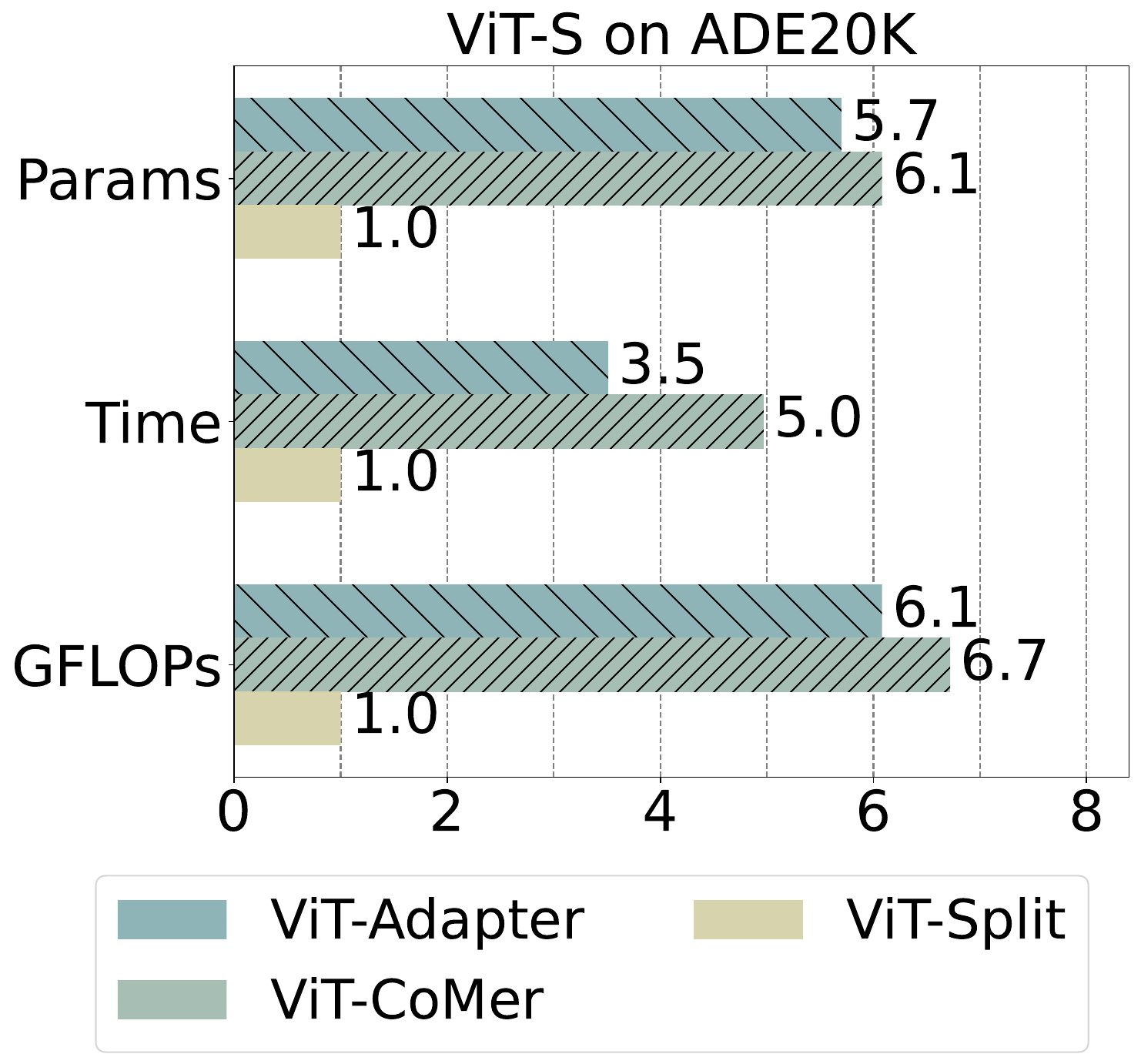}
        \caption{}\label{fig:vits_time}
    \end{subfigure}\hfill
    \begin{subfigure}[b]{0.23\textwidth}
        \centering
        \includegraphics[width=1\textwidth]{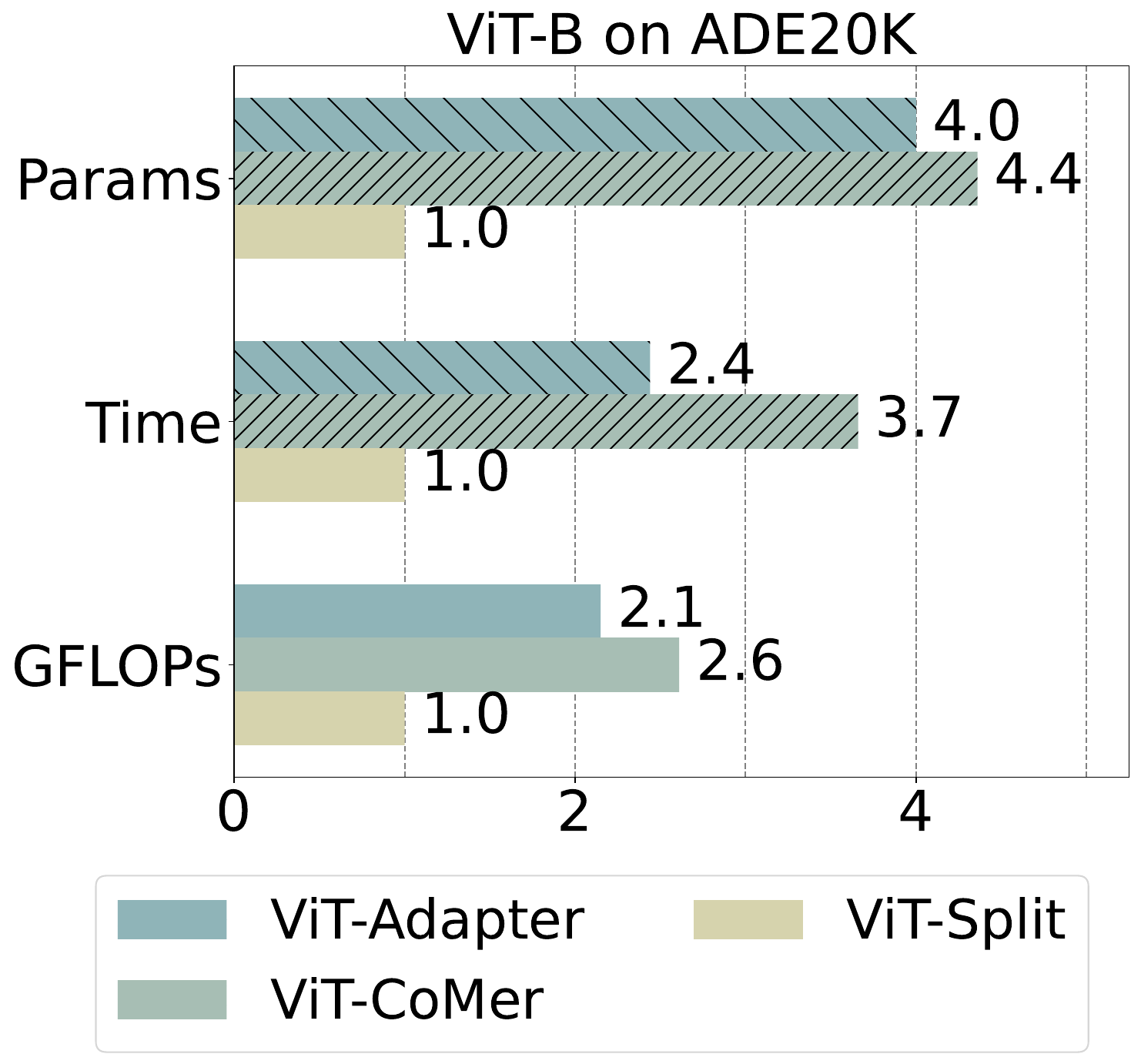}
        \caption{}\label{fig:vitb_time}
    \end{subfigure}\hfill
    \vspace{0.1cm}  
    \centering
    \scalebox{0.75}{    
    \begin{tabular}{cccc}
        \toprule
        Types & Train Params & Train Time (10,000 iters) & GFLOPS\\
        \midrule
        ViT-Split-S   & 10.1M   & 9m25s   & 129.9\\
        ViT-Split-B   & 33.2M   & 17m41s   & 508.6\\
        \bottomrule
    \end{tabular}
    }
    \label{tab:example_table}
    \caption{Comparison of time complexity for VFM adapters on ADE20K using two different sizes of ViT: (a) ViT-S and (b) ViT-B. For a fair evaluation, we reimplemented the other adapters under the same conditions, \ie, 4×A6000 Ada, over 10,000 iterations.}
    \label{fig:time_comparison}
\end{figure}

\textbf{ADE20K and Cityscapes val with 896$\times$896 image.} Additionally, we also compare with other SOTA methods on ADE20K (\cref{tab:seg_ADE20K_896}) and Cityscapes (\cref{tab:seg_cityscapes}) using images of 896$\times$896 resolution. As shown in \cref{tab:seg_ADE20K_896}, we can see that ViT-Split achieves results comparable to current SOTA methods on ADE20K val. It is worth mentioning that ViT-Split uses only a small linear head and does not rely on extra pretraining data. For a fair comparison, we benchmark against ViT-Adapter-G, which trains only the adapter and the Mask2former head based on the DINOv2 backbone. Our ViT-Split not only delivers better performance but also requires half the training parameters and achieves faster training speed. Specifically, according to \cite{oquabdinov2}, training ViT-Adapter-G requires 16 V100 GPUs for 28 hours, whereas our ViT-Split-G takes only 8 A6000 Ada GPUs for 15.7 hours. Moreover, on Cityscapes dataset (\cref{tab:seg_cityscapes}), our ViT-Split outperforms ViT-Adapter with only around 1/6 parameters being tuned. The results suggest that \emph{a simple linear head is enough for competitive results on semantic segmentation by fully leveraging VFM prior knowledge}.

\begin{table}[t]
\centering 
\small
\setlength{\tabcolsep}{2.5pt} 
\scalebox{0.84}{
    \begin{tabular}{l | ccccc}
    \hline
    {Method} &  Head    & \#Train Param     & mIoU (SS/MS)    & Iters     \\
    \hline
    DINOv2-B\textsuperscript{$\ddagger$}~\cite{oquabdinov2} & Linear & 127.0M & 81.2/82.3 & 20k \\
    \rowcolor[HTML]{E2F0D9}
    ViT-Split-B\textsuperscript{$\ddagger$} (ours) & Linear & 55.2M & 84.2/85.2 & 20k \\
    \hline
    Swin-L \cite{liu2021swin} & Oneformer \cite{jain2023oneformer} & 219M & 83.0/84.4 & 90k\\
    ViT-Adapter-L\textsuperscript{$\dagger$}~\cite{chenvision}                     & Mask2former  & 571M       & \underline{84.9}/\underline{85.8}     & 80k    \\
    DINOv2-L\textsuperscript{$\ddagger$}~\cite{oquabdinov2} & Linear & 312.9M & 83.5/84.3 & 20k \\
    \rowcolor[HTML]{E2F0D9}
    ViT-Split-L\textsuperscript{$\ddagger$} (ours) & Linear & 164.1M & \textbf{85.8}/\textbf{86.7} & 20k \\
    \hline
    \end{tabular}
}
\caption{\textbf{Semantic segmentation results on  Cityscales val with 896*896 resolution image. } ``{$\dagger$}’’ indicates that the model is initialized with BEiTv2 then pretrained on the Mapillary dataset. ``$\ddagger$'' represents the use of DINOv2. ``SS'' denotes single-scale testing, and ``MS'' means multi-scale testing.}
\label{tab:seg_cityscapes}
\end{table}

\begin{table*}[t!]
\centering
\setlength{\tabcolsep}{3.5pt} 
\scalebox{0.8}{
\begin{tabular}{l l c c c | c c c c c c c c c}
\toprule
\multirow{2}{*}{Method} & \multirow{2}{*}{LLM} & Image & \multicolumn{2}{c|}{Sample Size} & VQAv2 & VizWiz & LLaVA- & SciQA- & MM-Vet & \multicolumn{3}{c}{POPE \cite{li2023pope}} & MMB \\
 & & Size & Pre & Ft & \cite{goyal2017vqav2} & \cite{gurari2018vizwiz} & Wild~\cite{liu2024visual} & IMG~\cite{lu2022learn} & \cite{yu2023mm} & rand & pop & adv & \cite{liu2025mmbench} \\
\midrule
BLIP-2~\cite{li2023blip} & Vicuna-13B & 224$^2$ & 129M & - & 65.0 & 19.6 & 19.6 & 61 & 22.4 & \textbf{89.6} & 85.5 & 80.9 & --\\
InstructBLIP~\cite{dai2023instructblip} & Vicuna-7B & 224$^2$ & 129M & 1.2M & -- & 34.5 & 34.5 & 60.5 & 26.2 & -- & -- & -- & 36\\
InstructBLIP~\cite{dai2023instructblip} & Vicuna-13B & 224$^2$ & 129M & 1.2M & -- & 33.4 & 33.4 & 63.1 & 25.6 & 87.7 & 77 & 72 & --\\
Shikra~\cite{chen2023shikra} & Vicuna-13B & 224$^2$ & 600K & 5.5M & 77.4$^*$ & -- & -- & -- & -- & -- & -- & -- & 58.8\\
IDEFICS-9B~\cite{idefics} & LLaMA-7B & 224$^2$ & 353M & 1M & 50.9 & 35.5 & 35.5 & -- & -- & -- & -- & -- & 48.2\\
IDEFICS-80B~\cite{idefics} & LLaMA-65B & 224$^2$ & 353M & 1M & 60.0 & 36 & 36.0 & -- & -- & -- & -- & -- & 54.5\\
Qwen-VL~\cite{bai2023qwen} & Qwen-7B & 448$^2$ & 1.4B & 50M & \textbf{78.8}$^*$ & 35.2 & 35.2 & 67.1 & -- & -- & -- & -- & 38.2\\
Qwen-VL-Chat~\cite{bai2023qwen} & Qwen-7B & 448$^2$ & 1.4B$^*$ & 50M & 78.2$^*$ & 38.9 & 38.9 & \underline{68.2} & -- & -- & -- & -- & 60.6 \\
\midrule
LLaVA-1.5 \cite{liu2024improved} & Vicuna-7B & 336$^2$ & \textbf{558K} & \textbf{665K} & \underline{78.5}$^*$ & \underline{50.0}$^*$ & \underline{65.4} & 66.8 & \underline{31.1} & 87.3 & \underline{86.2} & \underline{84.2} & \underline{64.3} \\
\rowcolor[HTML]{E2F0D9}
LLaVA-1.5 + ViT-Split & Vicuna-7B & 336$^2$ & \textbf{558K} & \textbf{665K} & 78.2\textsubscript{\textbf{\textcolor{orange}{-0.3}}} & \textbf{51.7}\textsubscript{\textcolor{mygreen}{\textbf{+1.7}}} & \textbf{71.1}\textsubscript{\textcolor{mygreen}{\textbf{+5.7}}} & \textbf{70.4 }\textsubscript{\textcolor{mygreen}{\textbf{+3.6}}}& \textbf{31.2}\textsubscript{\textcolor{mygreen}{\textbf{+0.1}}} & \underline{88.5}\textsubscript{\textcolor{mygreen}{\textbf{+1.2}}} & \textbf{87.4}\textsubscript{\textcolor{mygreen}{\textbf{+1.2}}} & \textbf{86.1}\textsubscript{\textcolor{mygreen}{\textbf{+1.9}}} & \textbf{66.4}\textsubscript{\textcolor{mygreen}{\textbf{+2.1}}} \\
\bottomrule
\end{tabular}
}
\caption{\textbf{Comparison with different VLLM methods on VQA benchmarks.} ViT-Split is integrated into the vision encoder (CLIP-L) of LLaVA-1.5 (7B), tuning the penultimate block and utilizing prior feature from this layer. This adaptation can consistently enhance performance across most benchmarks, demonstrating the effectiveness and generalization of ViT-Split.  }
\label{tab:vqa_results_llava}
\end{table*}

\textbf{Time complexity analysis}. As illustrated in \cref{fig:time_comparison}, our ViT-Split achieves, on average, approximately 4× faster training speed for the small model and 3× faster for the base model compared to the other two VFM adapters. The slower training speed of the other adapters can be attributed to two factors: the early gradient backpropagation and the interaction between the CNN branch and the ViT. In contrast, our ViT-Split avoids backpropagating gradients to early layers, and reduces both the CNN branch computations and interaction overhead by fully leveraging the prior knowledge in the VFM. 
As shown in \cref{fig:inference_comparison}, traditional VFM adapters require training a task-specific VFM along with its corresponding adapter and head. In contrast, ViT-Split keeps the entire VFM frozen, training only a smaller adapter and the corresponding head. This design significantly reduces computational costs, making it more efficient for supporting multiple downstream tasks during inference.

\subsection{Detection and Instance Segmentation}

\textbf{Settings}. We present detection and instance segmentation results on COCO-2017 \cite{lin2014microsoft} in \cref{tab:detection}, using MMDetection~\cite{chen2019mmdetection}. The AdamW optimizer is employed with an initial learning rate of 1e-4 and a weight decay of 5e-2, training for 12 epochs (1$\times$ schedule). The total batch size is set to 16 and we utilize a MaskRCNN \cite{he2017mask} head for experiment. The setting of $K_p$ and $K_t$ is given in the Appendix.

As shown in \cref{tab:detection}, our ViT-Split achieves comparable performance with current SOTA VFM adapter ViT-CoMer. As discussed in \ref{observation}, the detection task may differ significantly from the original DINOv2 pretraining task, necessitating the tuning of more parameters. Despite this, our ViT-Split still involves \emph{fewer parameters and faster training speed (reducing 42\% training time) than ViT-CoMer}, demonstrating the efficiency of our architecture.

\begin{table}[t]
\centering
\small
\setlength{\tabcolsep}{2pt} 
\scalebox{0.92}{
\begin{tabular}{l | c | cccccc }
\hline
\multirow{2}{*}{Method} &
  \multirow{2}{*}{\#Param} &
  \multicolumn{6}{c}{Mask R-CNN 1× schedule} \\
  & & $ \mathrm{ AP^{b} }$ &
  $ \mathrm{ AP^{b}_{50} }$ &
  $ \mathrm{ AP^{b}_{75} }$ &
  $ \mathrm{ AP^{m} }$ &
  $ \mathrm{ AP^{m}_{50} }$ &
  $ \mathrm{ AP^{b}_{75} }$ \\ \hline
ConvNeXt-T~\cite{liu2022convnext}    & 48M  & 44.2 & 66.6 & 48.3 & 40.1 & 63.3 & 42.8 \\
Focal-T~\cite{yang2021focal}       & 49M  & 44.8 & 67.7 & 49.2 & 41.0 & 64.7 & 44.2 \\
SPANet-S~\cite{yun2023spanet} & 48M  & 44.7 & 65.7 & 48.8 & 40.6 & 62.9 & 43.8 \\
MixFormer-B4~\cite{chen2022mixformer}  & 53M  & 45.1  & 67.1 & 49.2 & 41.2 & 64.3 & 44.1 \\
Twins-B~\cite{chu2021twins} & 76M  & 45.2 & 67.6 & 49.3 & 41.5 & 64.5 & 44.8 \\
Swin-S~\cite{liu2021swin} & 69M  & 44.8 & 66.6 & 48.9 & 40.9 & 63.4 & 44.2 \\
Flatten-PVT-T~\cite{han2023flatten} & 49M  & 44.2 & 67.3 & 48.5 & 40.2 & 63.8 & 43.0 \\
ViT-S~\cite{li2021benchmarking}         & 44M  & 40.2 & 63.1 & 43.4 & 37.1 & 60.0 & 38.8 \\
ViTDet-S~\cite{li2022vitdet}      & 46M  & 40.6 & 63.3 & 43.5 & 37.1 & 60.0 & 38.8 \\
ViT-Adapter-S~\cite{chenvision} & 48M  & 44.7 & 65.8 & 48.3 & 39.9 & 62.5 & 42.8 \\
ViT-CoMer-S~\cite{xia2024vit} &  50M  &  45.8  & 67.0 & 49.8 & 40.5 & 63.8 & 43.3 \\
ViT-CoMer-S{$\ddagger$}~\cite{xia2024vit} &  50M  &  \textbf{48.6}  & \textbf{70.5} & \underline{53.1} & \textbf{42.9} & \underline{67.0} & \textbf{45.8} \\ 
\rowcolor[HTML]{E2F0D9}
ViT-Split-S\textsuperscript{$\ddagger$} (ours)  &  45M  &  \underline{48.5}  & \textbf{70.5} & \textbf{53.3} & \underline{42.8} & \textbf{67.2} & \underline{45.6} \\ 

\hline
PVTv2-B5~\cite{wang2022pvtv2}     & 102M & 47.4 & 68.6 & 51.9 & 42.5 & 65.7 & 46.0 \\
InternImage-B~\cite{wang2023internimage} & 115M & 48.8 & 70.9 & 54.0 & 44.0 & 67.8 & 47.4 \\
ViT-B~\cite{li2021benchmarking}         & 114M & 42.9 & 65.7 & 46.8 & 39.4 & 62.6 & 42.0 \\
ViTDet-B~\cite{li2022vitdet}      & 121M & 43.2 & 65.8 & 46.9 & 39.2 & 62.7 & 41.4 \\
ViT-Adapter-B~\cite{chenvision} & 120M & 47.0 & 68.2 & 51.4 & 41.8 & 65.1 & 44.9 \\
ViT-CoMer-B \cite{xia2024vit}        &  129M   & 47.6 & 68.9 & 51.9 & 41.8 & 65.9 & 44.9 \\ 
ViT-CoMer-B\textsuperscript{$\ddagger$} \cite{xia2024vit}        &  129M   & \textbf{52.0} & \textbf{73.6} & \textbf{57.2} & \textbf{45.5} & \textbf{70.6} & \textbf{49.0} \\ 
\rowcolor[HTML]{E2F0D9}
ViT-Split-B\textsuperscript{$\ddagger$} (ours)  &  118M  &  \underline{51.8}  & \textbf{73.6} & \underline{57.1} & \underline{45.4} & \underline{70.3} & 48.6 \\ 
\hline
ViT-L\textsuperscript{$\dagger$}~\cite{li2021benchmarking}         & 337M & 45.7 & 68.9 & 49.4 & 41.5 & 65.6 & 44.6 \\
ViTDet-L\textsuperscript{$\dagger$}~\cite{li2022vitdet}      & 351M & 46.2 & 69.2 & 50.3 & 41.4 & 65.8 & 44.1 \\
ViT-Adapter-L\textsuperscript{$\dagger$}~\cite{chenvision} & 348M & 48.7 & 70.1 & 53.2 & 43.3 & 67.0 & 46.9 \\
ViT-CoMer-L\textsuperscript{$\dagger$} \cite{xia2024vit}        &  363M & 51.4 & 73.5 & 55.7 & 45.2 & 70.3 & 48.5 \\ 
ViT-CoMer-L\textsuperscript{$\ddagger$} \cite{xia2024vit}        &  363M & \textbf{53.4} & \textbf{75.3} & \textbf{58.9} & \textbf{46.8} & \textbf{72.0} & \textbf{50.9} \\ 
\rowcolor[HTML]{E2F0D9}
ViT-Split-L\textsuperscript{$\ddagger$} (ours)  &  348M  &  \underline{53.0}  & \underline{75.1} & \underline{58.1} & \underline{46.6} & \underline{71.9} & \underline{50.4} \\ 
\hline
\end{tabular}
}
\caption{\textbf{Object detection and instance segmentation using Mask R-CNN on COCO val2017.} ``$\dagger$'' indicates pre-training with ImageNet-22K, $\ddagger$'' represents the use of DINOv2~\cite{oquabdinov2}, while the default setting uses ImageNet-1K pre-training.}
\label{tab:detection}
\end{table}

\subsection{Visual Question Answering}

\textbf{Settings}. We also present VQA results using the popular visual large language model (VLLM) \cite{li2025visual}, LLaVA-1.5 \cite{liu2024improved}. This model comprises a CLIP-L visual encoder for encoding images, an MLP connector for projecting visual tokens into the language space, and a Vicuna-based LLM \cite{chiang2023vicuna} for generating language tokens.  In our modified LLaVA, we replace the original MLP projector with our ViT-Split.  To comprehensively evaluate the effectiveness of our ViT-Split, we utilize both academic-task-oriented benchmarks ( VQA-v2 \cite{goyal2017vqav2}, VizWiz \cite{gurari2018vizwiz}, SciQA-IMG \cite{lu2022learn}), and instruction-following LLM benchmarks (POPE \cite{li2023pope}, MMBench \cite{liu2025mmbench}, LLaVA-Wild \cite{liu2024visual}, MM-Vet~\cite{yu2023mm}).  Following \cite{liu2024improved}, we first pretrain our ViT-Split using 558K image-text pairs, and subsequently fine-tune both ViT-Split and the LLM with 665K mixed data pairs. For more detailed information regarding the hyperparameter settings, please refer to the Appendix.

As shown in \cref{tab:vqa_results_llava}, our ViT-Split enhances LLaVA-1.5 performance across most benchmarks. This improvement demonstrates that ViT-Split is also  applicable to other VFMs and VQA tasks. Unlike most current VLLMs that directly utilize features from the penultimate layer, ViT-Split leverages both the prior features of the vision encoder and the task-specific features, resulting in richer visual representations that improve the LLM's learning process. Moreover, we tune only a small portion of the vision encoder's parameters (specifically, one layer), which ensures efficiency for both training and inference. We believe that ViT-Split will offer new inspiration for VLLM design.

\subsection{Ablation Study}
We conduct an ablation study for each trainable component in \cref{tab:ablation} on ADE20K. The default settings are consistent with those described in \cref{sec:seg}.

\textbf{The effectiveness of prior head}. The results in \cref{tab:ablation} show that incorporating the prior head improves performance by 2.7\% and 3.6\% compared to the baseline that uses only the final-layer features. This suggests that the prior head effectively leverages multi-layer prior features from the VFM to enhance overall representation quality, surpassing the use of solely the final layer’s prior features. Additionally, our module enhances 2D local representations through the use of a CNN. Furthermore, the results demonstrate that the prior features extracted from the original VFM are highly valuable, achieving performance levels nearly equivalent to those obtained through full fine-tuning.

\textbf{The effectiveness of task head}. As shown in \cref{tab:ablation}, by tuning only the task head $g_{\theta_t}$, the performance nearly matches that of fine-tuning the entire model, supporting the finding in \cref{observation}. Last few layers can learn task-specific features and achieve similar performance as tuning the entire backbone. Furthermore, the experiments demonstrate that performance can be further enhanced when combined with prior features. We attribute this improvement to the combined benefits of task-specific and prior knowledge, with the latter helping to reduce task head overfitting.

\textbf{The effectiveness of fusion head}. \cref{tab:ablation} shows that using fusion net $g_{\theta_f}$ yields a performance improvement of 1.1\% for two ViT sizes. We attribute this enhancement to our CNN-based fusion module, which retains richer feature information compared to a simple addition operation. Again, the CNN component strengthens the local feature representation, contributing to  improved fusion results.
\begin{table}[t]
    \centering
    \setlength{\tabcolsep}{3.5pt} 
    \scalebox{0.88}{    
    \begin{tabular}{ccccccc}
        \toprule
        \multicolumn{3}{c}{Components} & \multicolumn{2}{c}{Train Params} & \multicolumn{2}{c}{mIoU}\\
        prior $g_{\theta_p}$ & task $g_{\theta_t}$ & fusion $g_{\theta_f}$ & Small & Base & Small & Base \\
        \midrule
        \rowcolor[HTML]{EBEBEB}
          & & & 22.1M  & 91.4M & 49.6 &  53.8 \\
          & & & 1.2M  & 4.84M & 44.3 &  47.8 \\
         \checkmark & & & 3.2M  & 12.6M & 46.0 &  51.4 \\
        & \checkmark &  & 6.6M  & 26.1M  & 49.5 & 53.2 \\
        \checkmark & \checkmark &  & 8.6M   & 33.9M & 50.4 & 54.6  \\
        \rowcolor[HTML]{E2F0D9}
        \checkmark & \checkmark & \checkmark  & 10.2M   & 40.5M & 51.6 & 55.7  \\
        \bottomrule
    \end{tabular}
    }
    \caption{Ablation study of the prior head ($g_{\theta_p}$), task head ($g_{\theta_t}$), and fusion net ($g_{\theta_f}$) on ADE20K, conducted with two ViT sizes: small and base on ViT-Split\textsubscript{u}. We set $K_t = 3$ and $K_p = 4$ for both model sizes. The baseline model (no modules used, shown without background color) uses only the frozen features from the last layer. The baseline with a gray background indicates full fine-tuning of the entire backbone. When only $g_{\theta_p}$ and $g_{\theta_t}$ are used, their features are combined via addition.}
    \label{tab:ablation}
\end{table}

\begin{table}[t]
    \centering
    \setlength{\tabcolsep}{4.5pt} 
    \scalebox{0.85}{    
    \begin{tabular}{ccccc}
        \toprule
        \multirow{2}{*}{Select strategy} &  \multicolumn{3}{c}{mIoU}\\
         & Small & Base & Large\\
        \midrule
        ${\rm{last\, few\, layers}}$  & 51.3 & 54.9 & 57.3\\
        \rowcolor[HTML]{E2F0D9}
        ${\rm{uniform}}$   & 51.6 & 55.7 & 58.2\\
        \bottomrule
    \end{tabular}
    }
    \caption{Ablation study on the frozen layer selection strategies for our ViT-Split model on the ADE20K dataset, using three ViT sizes: small, base, and large. $K_p$ is same for all  strategies.}
    \label{tab:ablation_selected_layers}
\end{table}

\textbf{The effectiveness of uniform layer selection.} In \cref{tab:ablation_selected_layers}, we evaluate the effectiveness of the selection strategy for prior features. Compared to selecting features from only the last few layers, which capture mostly task-specific prior information—uniform selection allows for a more diverse set of prior features, encompassing both low-level and task-specific characteristics. This uniform selection approach becomes increasingly impactful as the backbone size grows.

\begin{figure}
    \centering
    \begin{subfigure}[b]{0.235\textwidth}
        \centering
        \includegraphics[width=1\textwidth]{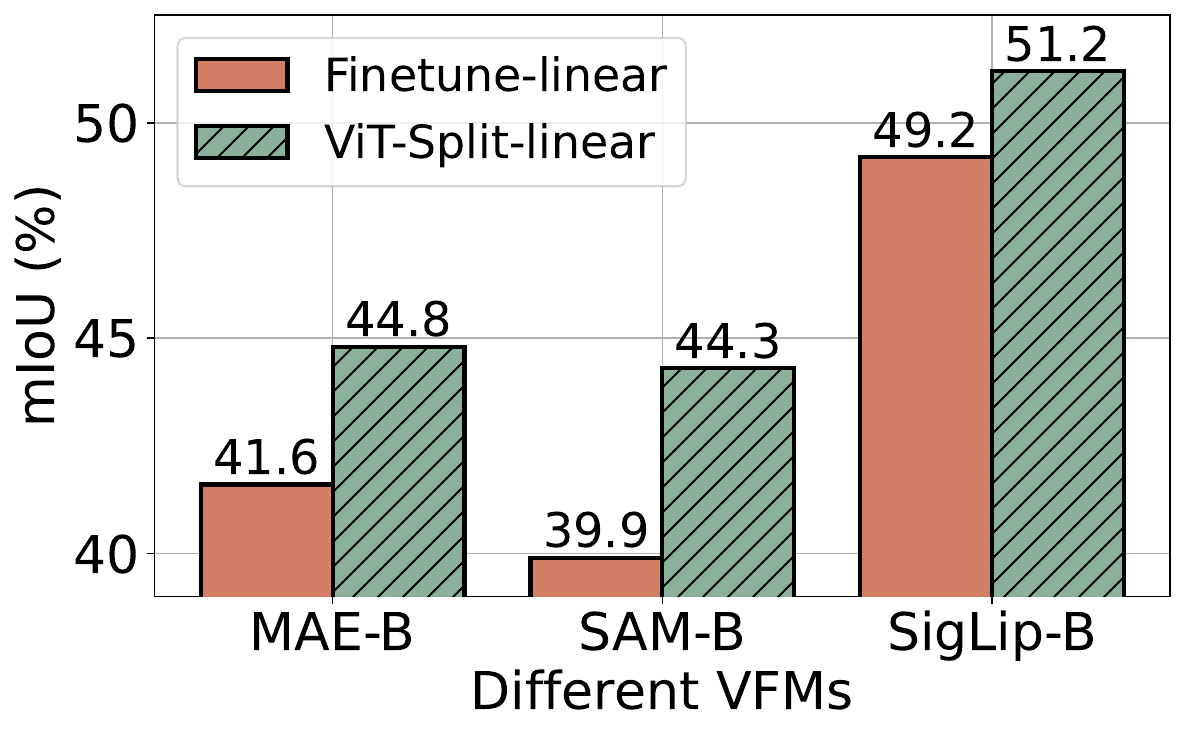}
        \caption{mIoU.}
    \end{subfigure}\hfill
    \begin{subfigure}[b]{0.235\textwidth}
        \centering
        \includegraphics[width=1\textwidth]{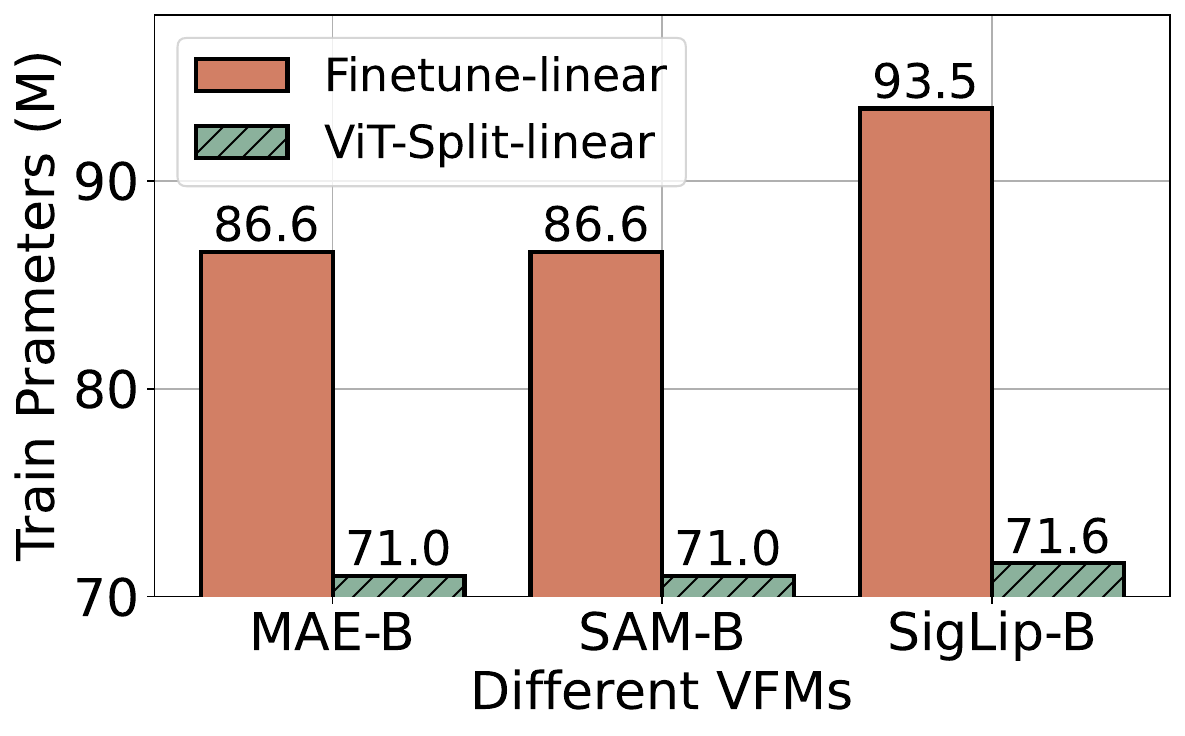}
        \caption{Training parameters.}
    \end{subfigure}
    \caption{Segmentation results and parameters on ADE20K with different VFMs, including MAE-B \cite{he2022masked}, SAM-B \cite{kirillov2023segment} and SigLip-B~\cite{zhai2023sigmoid}. We set $K_p=4$ and $K_t=8$ for all the VFMs.}
    \label{fig:different_VFMs_seg}
\end{figure}

\textbf{The effectiveness across different VFMs.}  To evaluate the generality of our ViT-Split, we present results on various VFMs in \cref{fig:different_VFMs_seg}, leveraging the excellent VFM-benchmark codebase~\footnote{\url{https://github.com/tue-mps/benchmark-vfm-ss}}. The experiments demonstrate that ViT-Split consistently enhances performance across both weakly-supervised VFMs (SAM and SigLip) and self-supervised VFMs (MAE). These results not only validate the effectiveness of ViT-Split on multiple VFMs but also suggest that our observations may hold for a broader range of VFMs.

\section{Conclusion}

In this paper, we introduce ViT-Split, an efficient, effective, and generalized adapter, to adapt VFMs for downstream tasks. Specifically, we introduce two heads based on a frozen VFM, a prior head for multi-scale prior feature extraction and a task head for task-specific feature adaptation. Experiments on segmentation, detection, MDE, and VQA verify the effectiveness and efficiency of our method. In the future, we aim to apply ViT-Split to more VFMs and tasks. We hope our method offers a fresh perspective for efficient and effective VFM adapter design.

\clearpage
\setcounter{page}{1}
\appendix
\maketitlesupplementary
\begin{figure}
    \centering
    \includegraphics[width=0.9\linewidth]{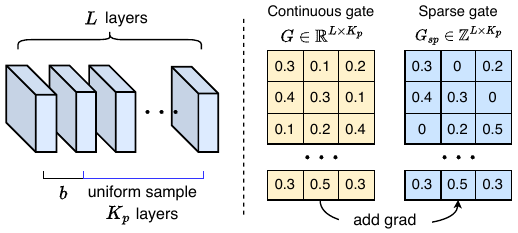}
    \caption{Illustration of our proposed layer selection methods: uniform sampling (left) and sparse gate (right). Uniform sampling selects $K_p$ layers from $L$ prior features, ranging from the $b$-th to $L$-th layer. The sparse gate, utilizing the STE technique (see \cref{eq:sparse_gate_eq}), aggregates multiple layer features and filters out irrelevant ones.}
    \label{fig:layer-selection}
\end{figure}
\section{Training details}
\label{sec:training_details}
\begin{table}[t]
\centering 
\small
\setlength{\tabcolsep}{3pt} 
\scalebox{1}{
    \begin{tabular}{l | ccccc}
    \hline
    {Method} &  Head    & \#Train Param     & mIoU   & Iters     \\
    \hline
    ViT-Split-L (sparse gate)  & Linear & 164.1M & \underline{85.7} & 20k \\
    ViT-Split-L (uniform)  & Linear & 164.1M & \textbf{85.8} & 20k \\
    \hline
    \end{tabular}
}
\caption{{Comparison of two layer selection methods on semantic segmentation. The results are conducted on  Cityscales val with 896*896 resolution image. } }
\label{tab:selection_seg_cityscapes}
\end{table}

\begin{table}[t]
\centering 
\small
\setlength{\tabcolsep}{3pt} 
\scalebox{1}{
    \begin{tabular}{l | ccccc}
    \hline
    {Method} &  Head    & \#Train Param     & mIoU   & Iters     \\
    \hline
    ViT-Split-S (sparse gate) & Linear & 10.2M & \underline{51.5} & 40k \\
    ViT-Split-S (uniform) & Linear & 10.2M & \textbf{51.6} & 40k \\
    \hline
    ViT-Split-B (sparse gate) & Linear & 40.5M & \underline{55.5} & 40k \\
    ViT-Split-B (uniform)  & Linear & 40.5M & \textbf{55.7} & 40k \\
    \hline
    ViT-Split-L (sparse gate) & Linear & 88.6M & \underline{58.1} & 40k \\
    ViT-Split-L (uniform)  & Linear & 88.6M & \textbf{58.2} & 40k \\
    \hline
    \end{tabular}
}
\caption{{Comparison of two layer selection methods on semantic segmentation. The results are conducted on  ADE20K val with 512*512 resolution image. } }
\label{tab:selection_seg_ade20k}
\end{table}

\subsection{Hyper-parameter setting}
We outline the settings for several key hyperparameters of ViT-Split in \cref{tab:hyper_params}, including weight initialization, the number of tuning layers ($K_t$), and the number of selected prior features ($K_p$), \etc. We conduct experiments across four tasks: semantic segmentation, monocular depth prediction, detection, and visual question answering (VQA).

\textbf{The selection guideline of $K_t$, $K_p$ and $b$.} As shown in \cref{tab:hyper_params}, these hyperparameters vary across tasks, with their importance ranked as $K_t>K_p>b$. As shown in \cref{fig:param_sensitivity}, $K_t$ is the most critical hyperparameter and is task-dependent. For dense prediction tasks (\textit{e.g.}, segmentation or monocular depth estimation), tuning smaller layers (around $1/6$ to $1/4$) yields good performance. For detection tasks, since the pretrained task differs significantly from detection (see Fig. 4), tuning more layers is necessary for better results. $K_p$ has a smaller impact on results compared to $K_t$, and $K_p=4$ works well in most cases. Typically, we set $b=2$ to sample prior features from both shallow and deep layers. However, for tasks like VQA, only the last-layer features are needed, as the LLM decoder benefits more from high-level features while low-level features may introduce noise. 

\begin{table*}[t]
    \centering
    \setlength{\tabcolsep}{3pt} 
    \scalebox{0.9}{    
    \begin{tabular}{cccccccc}
        \toprule
        Method & Initialization & Tasks & Datasets & Image res. & $K_t$ & $K_p$ & $b$\\
        \midrule
        ViT-Split-S & DINOv2 & Semantic segmentation & ADE20K & (512, 512) & 3 & 4 & 2 \\
        ViT-Split-B & DINOv2 & Semantic segmentation & ADE20K & (512, 512) & 3 & 4 & 2 \\
        ViT-Split-L & DINOv2 & Semantic segmentation & ADE20K & (512, 512) & 4 & 8 & 2 \\
        ViT-Split-L & DINOv2 & Semantic segmentation & ADE20K & (896, 896) & 4 & 8 & 2 \\
        ViT-Split-G & DINOv2 & Semantic segmentation & ADE20K & (896, 896) & 8 & 14 & 26 \\
        ViT-Split-L & DINOv2 & Semantic segmentation & Cityscapes & (896, 896) & 10 & 8 & 2 \\
        ViT-Split-L & DINOv2 & Semantic segmentation & Pascal Context & (480, 480) & 6 & 10 & 2 \\
        ViT-Split-S & DINOv2 & Monocular depth estimation & NYU-V2 & (416, 544) & 3 & 4 & 2 \\
        ViT-Split-B & DINOv2 & Monocular depth estimation & NYU-V2 & (416, 544) & 4 & 4 & 2 \\
        ViT-Split-L & DINOv2 & Monocular depth estimation & NYU-V2 & (416, 544) & 3 & 4 & 6 \\
        ViT-Split-S & DINOv2 & Detection, instance segmentation & COCO17 & (1024, 1024) & 11 & 4 & 2 \\
        ViT-Split-B & DINOv2 & Detection, instance segmentation & COCO17 & (1024, 1024) & 11 & 6 & 2 \\
        ViT-Split-L & DINOv2 & Detection, instance segmentation & COCO17 & (1024, 1024) & 23 & 8 & 6 \\
        LLaVA+ViT-Split-L & CLIP & Vision question answering & VQA benchmarks & (512, 512) & 1 & 1 & 23 \\
        \bottomrule
    \end{tabular}
    }
    \caption{The settings of the important hyper-parameters of ViT-Split on different tasks, including semantic segmentation, monocular depth estimation, detection and instance segmentation, and vision question answering (VQA).}
    \label{tab:hyper_params}
\end{table*}

\begin{table*}[t]
\centering 
\small
\setlength{\tabcolsep}{4.8pt} 
\scalebox{0.95}{
    \begin{tabular}{l | cccccccc}
    \hline
    {Architecture} &  Head    & \#Train Param ($\downarrow$)     & AbsRel ($\downarrow$)    & RMSE ($\downarrow$) &  $\log_{10}$ ($\downarrow$) & $\delta_1$ ($\uparrow$) & $\delta_2$ ($\uparrow$) & $\delta_3$ ($\uparrow$)\\
    \hline
    ResNet-101 \cite{he2016deep}  & DORN \cite{fu2018deep} & 110M & 0.115 & 0.509 & 0.051 & 0.828 & 0.965 & 0.992 \\
    ViT \cite{dosovitskiy2020image}  & DPT \cite{ranftl2021vision} & -- & 0.110 & 0.357 & 0.045 & 0.904 & 0.988 & 0.998 \\
    EfficientNet-B5 \cite{tan2019efficientnet}  & AdaBins \cite{bhat2021adabins} & 77M & 0.103 & 0.364 & 0.044 & 0.903 & 0.984 & 0.997 \\
    ResNet-101 \cite{he2016deep} & P3Depth \cite{patil2022p3depth} & -- & 0.104 & 0.356 & 0.043 & 0.898 & 0.981 & 0.996 \\
    Swin-L \cite{liu2021swin}  & BinsFormer \cite{li2024binsformer} & 273M & 0.094 & 0.329 & 0.04 & 0.923 & 0.989 & 0.997 \\
    Swin-L \cite{liu2021swin}  & NeWCRFs \cite{yuan2022neural} & 270M & 0.095 & 0.334 & 0.041 & 0.922 & 0.992 & -- \\
    Swin-L \cite{liu2022swin}  & PixelFormer \cite{agarwal2023attention} & -- & 0.090 & 0.322 & 0.039 & 0.929 & 0.991 & 0.998 \\
    Swin-L \cite{liu2021swin}  & VA-DepthNet \cite{liu2023vadepth} & -- & 0.086 & 0.304 & -- & 0.937 & 0.992 & 0.998 \\
    Swin-L \cite{liu2021swin}  & iDisc \cite{piccinelli2023idisc} & 209M & 0.086 & 0.314 & 0.038 & 0.940 & 0.993 & 0.999 \\
    Swin-L \cite{liu2022swin}  & IEBins \cite{shao2024iebins} & 273M & 0.087 & 0.314 & 0.038 & 0.936 & 0.992 & 0.998 \\
    SwinV2-L \cite{liu2022swin}  & AiT-P \cite{ning2023all} & -- & 0.076 & 0.275 & 0.033 & 0.954 & 0.994 & 0.999 \\
    \hline
    DINOV2-G\textsuperscript{$\ddagger$} \cite{oquabdinov2} & DPT \cite{ranftl2021vision} & -- & 0.0907 & 0.279 & 0.0371 & 0.9497 & 0.996 & 0.9994 \\
    \rowcolor[HTML]{E2F0D9}
    ViT-Split-S\textsuperscript{$\ddagger$} & Linear & 9.3M & 0.0897 & 0.3358 & 0.039 & 0.9327 & 0.9908 & 0.9985 \\
    \rowcolor[HTML]{E2F0D9}
    ViT-Split-B\textsuperscript{$\ddagger$} & Linear & 37.0M & 0.0853 & 0.3019 & 0.0365 & 0.9412 & 0.9947 & 0.9991 \\
    \rowcolor[HTML]{E2F0D9}
    ViT-Split-L\textsuperscript{$\ddagger$} & Linear & 65.5M & \textbf{0.078} &\textbf{ 0.2672} & \textbf{0.0327} & \textbf{0.9622} & \textbf{0.9967} & \textbf{0.9994} \\
    \hline
    \end{tabular}
}
\caption{\textbf{Monocular depth estimation results on  NYU-V2 with 416*544 resolution image. } ``$\ddagger$'' represents the use of DINOv2. Other backbones are initialized with ImageNet-1K/22K weights}
\label{tab:mde_NYU}
\end{table*}

\subsection{Sizes of various heads}
We provide the sizes of the various heads used in ViT-Split for different tasks in \cref{tab:different_heads}, including segmentation (seg.), detection (det.) and monocular depth estimation (mde).
\begin{table}[h]
    \centering
    \setlength{\tabcolsep}{4pt} 
    \scalebox{0.9}{    
    \begin{tabular}{cccccccc}
        \toprule
        Architectures & Task & Small & Base & Large \\
        \midrule
        Linear & segmentation & 4.78M & 19.00M & 37.91M \\
        Mask-RCNN & detection & 17.54M & 17.54M & 17.54M \\
        Linear & MDE & 0.34M & 1.28M & 2.23M \\
        \bottomrule
    \end{tabular}
    }
    \caption{The size of different heads used for ViT-Split.}
    \label{tab:different_heads}
\end{table}

\subsection{Details of tuning the VLLM}
LLaVA-1.5 employs a CLIP-based vision encoder for image encoding. We introduce a single-layer task head copied from CLIP’s original final layer  (\emph{i.e.}, $K_t=1$) and utilize only the last-layer feature of CLIP as the input to the prior head (\emph{i.e.}, $K_p=1$). We replace the original MLP projector in LLaVA-1.5 with with our ViT-Split for two-stage training. The training follows the same hyperparameter settings as the original LLaVA-1.5. 

\subsection{Architecture details of various used VFMs}
We provide the architecture details of various VFMs used in the main content in \cref{tab:different_VFMs}.
\begin{table}[h]
    \centering
    \setlength{\tabcolsep}{3pt} 
    \scalebox{0.9}{    
    \begin{tabular}{cccccccc}
        \toprule
        Architectures & Embed dim & Layers & Params (M) \\
        \midrule
        DINOv2-S & 384 & 12 & 22 \\
        DINOv2-B & 768 & 12 & 87 \\
        DINOv2-L & 1024 & 24 & 304 \\
        DINOv2-G & 1536 & 40 & 1137 \\
        CLIP-L & 1024 & 24 & 428 \\
        \bottomrule
    \end{tabular}
    }
    \caption{The architecture details of used VFMs.}
    \label{tab:different_VFMs}
\end{table}

\section{Layer selection}

\subsection{Sparse gate}
Another way is to learn the sparse gate $G_{sp}\in \mathbb{R}^{L\times K_p}$ from the dataset. This method eliminates the need for carefully tuning hyperparameters to select prior features. To remove noisy features, we enforce the sparsity in the gate by selecting top $K_p$ scores, and normalizing the remained ones. However, directly optimizing $G_{sp}$ is infeasible since the sparsity operation is non-differentiable. To address this issue, we employ the Straight-Through Estimator (STE) technique which allows for approximate gradient optimization. Specifically, let $G\in \mathbb{R}^{L\times K_p}$ be the learnable gate, which is continuous. From $G$, we obtain the sparse gates $G_{sp}$ by selecting the top $K_p$ elements in each column. We then apply STE by optimizing the gradient of $G$:
\begin{equation}\label{eq:sparse_gate_eq}
    G_{sp} = G_{sp} + G - G_{no\_grad}.
\end{equation}
After obtaining the sparse gate $G_{sp}\in \mathbb{R}^{L\times K_p}$, we can get the selected prior features by multiplying with the prior feature map $\mathbf{f}_p\in \mathbb{R}^{h\times w\times L \times D}$ from the layer dimension.

\subsection{Performance on segmentation task}

We present a comparison of layer selection methods on segmentation benchmarks, including Cityscapes and ADE20K, in \cref{tab:selection_seg_cityscapes} and \cref{tab:selection_seg_ade20k}. For a fair comparison, we set the same $K_t$ for both selection methods and use $K_p=4$ for all sparse-gate-based experiments. Our results show that sparse gate selection achieves comparable performance to uniform sampling on segmentation tasks without requiring manual hyper-parameter selection. It indicates that sparse gate selection is a promising and versatile approach for reducing the number of hyper-parameters.

\section{Motivation of freezing the backbone}
Freezing the backbone has three main motivations. \ding{172} \textbf{Improved training and inference speed.} Fig. 7  shows our ViT-Split achieves 2.4$\sim$5$\times$, and 2$\sim$6$\times$ speedup over other VFM-Adapters on training and inference efficiency. Additionally, as detailed in \cref{training_time_heads}, ViT-Split is 1.4$\sim$3$\times$ faster than finetuning the entire backbone with a linear/UperNet head. \ding{173} \textbf{Enhanced performance with prior features.}  We admit that the inference speed will decrease compared with finetuning DINOv2-linear due to the extra heads (around 30\% on segmentation tasks). However, the performance can be further improved, which is also the main motivation of other VFM-adapters. Compared with these, ViT-Split achieves better training and inference efficiency. \ding{174} \textbf{Task adaptivity.} ViT-Split requires storing only separate task-specific heads, rather than the entire model, making it more adaptive and memory-efficient for deployment across multiple tasks.

\section{Explanation of the lower performance on detection task}
We acknowledge that the performance difference between ViT-Split and ViT-CoMer on Mask R-CNN (Tab. 5) is relatively small. However, ViT-Split uses only 90\%–95\% of ViT-CoMer’s trainable parameters, already demonstrating clear advantages in training efficiency while maintaining comparable accuracy. The primary reason ViT-Split does not significantly outperform other VFM-adapters lies in the relatively weak task alignment of the prior features from DINOv2 for object detection tasks. Unlike DETR-style models, which are pre-trained with strong detection-oriented objectives, self-supervised models like DINOv2 tend to provide less directly transferable features for detection. This necessitates using more layers in the task head (\emph{i.e.}, larger $K_t$), effectively making ViT-Split rely more on fine-tuning, similar to other VFM-adapters. As self-supervised models begin to offer stronger detection-aware priors, we expect ViT-Split to better leverage them and close the gap with current SOTA DETR-style models.

\section{More results}
\subsection{An apple-to-apple comparison with other VFM-adapters on segmentation}
We provide an apple-to-apple comparison with the SOTA VFM-adapters in \cref{tab:vfm_adapter_comparison}, \textit{i.e.}, ViT-CoMer \cite{xia2024vit} and ViT-Adapter \cite{chenvision}. All models are trained for 40K iterations on ADE20K, using a UperNet head for the baselines and a linear head for ViT-Split. For VFM-adapters, we adopt a learning rate schedule similar to that used in detection tasks, incorporating layer-wise decay with carefully tuned rates for each baseline to ensure strong performance. Results show that with DINOv2 initialization, ViT-Split consistently outperforms other VFM-adapters across different model sizes. This highlights ViT-Split’s ability to better leverage the strong prior knowledge from DINOv2 without altering the original feature representations, which often results in suboptimal performance in other adapters.

\begin{table}[h]
    \centering
    \caption{VFM-adapter comparison on ADE20K (40K iterations).} \label{tab:vfm_adapter_comparison}
    \scalebox{0.75}{    
    \begin{tabular}{lcccccc}
    \toprule
    \multirow{2}{*}{\textbf{Method}} 
    & \multicolumn{2}{c}{\textbf{DINOv2-S}} 
    & \multicolumn{2}{c}{\textbf{DINOv2-B}} 
    & \multicolumn{2}{c}{\textbf{DINOv2-L}} \\
    \cmidrule(lr){2-3} \cmidrule(lr){4-5} \cmidrule(lr){6-7}
    & \textbf{mAP} & \textbf{\#Param}
    & \textbf{mAP} & \textbf{\#Param}
    & \textbf{mAP} & \textbf{\#Param} \\
    \midrule
    ViT-Adapter     & 45.4 & 57.1M  & 49.9 & 134.5M  & 52.2 & 364.6M \\
    ViT-CoMer   & 44.6 & 61.8M  & 48.4 & 145.6M  & 51.8 & 384.2M  \\
    ViT-Split   & 51.6 & 10.2M  & 55.7 & 40.5M  & 58.2 & 88.6M  \\
    \bottomrule
    \end{tabular}
    }
\end{table}

\subsection{Hyper-parameter sensitivity analysis}\label{sec:hyper-param_analysis}
\begin{figure}[t]
    \centering
    \begin{subfigure}[b]{0.23\textwidth}
        \centering
        \includegraphics[width=1\textwidth]{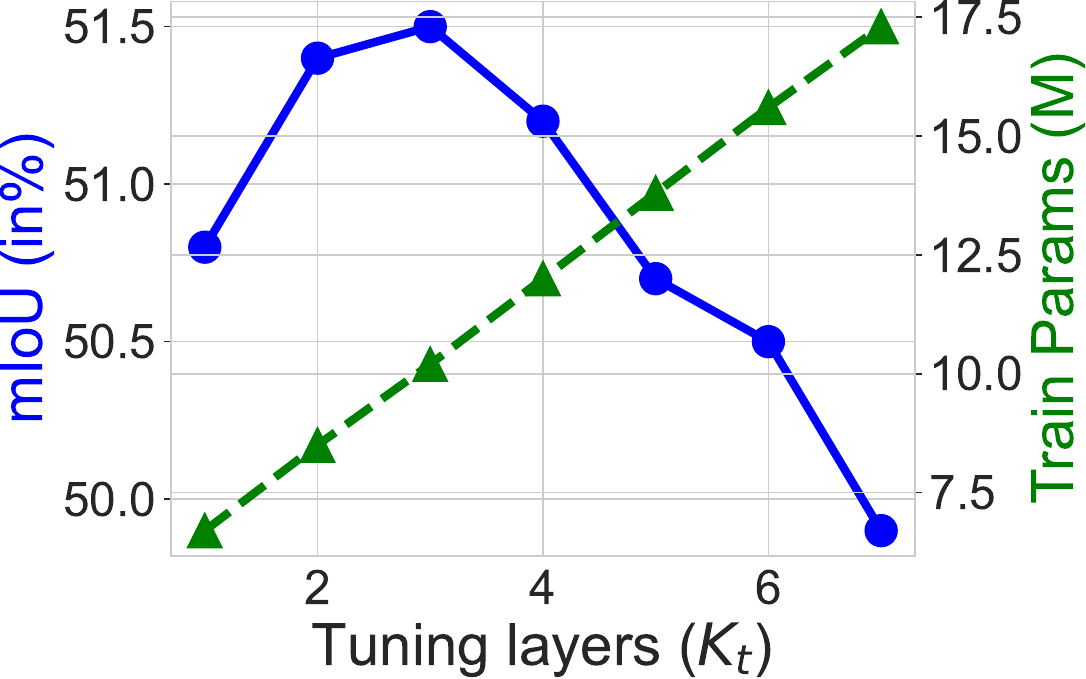}
        \caption{Tuning layers $K_t$.}\label{fig:Kt}
    \end{subfigure}\hfill
    \begin{subfigure}[b]{0.23\textwidth}
        \centering
        \includegraphics[width=1\textwidth]{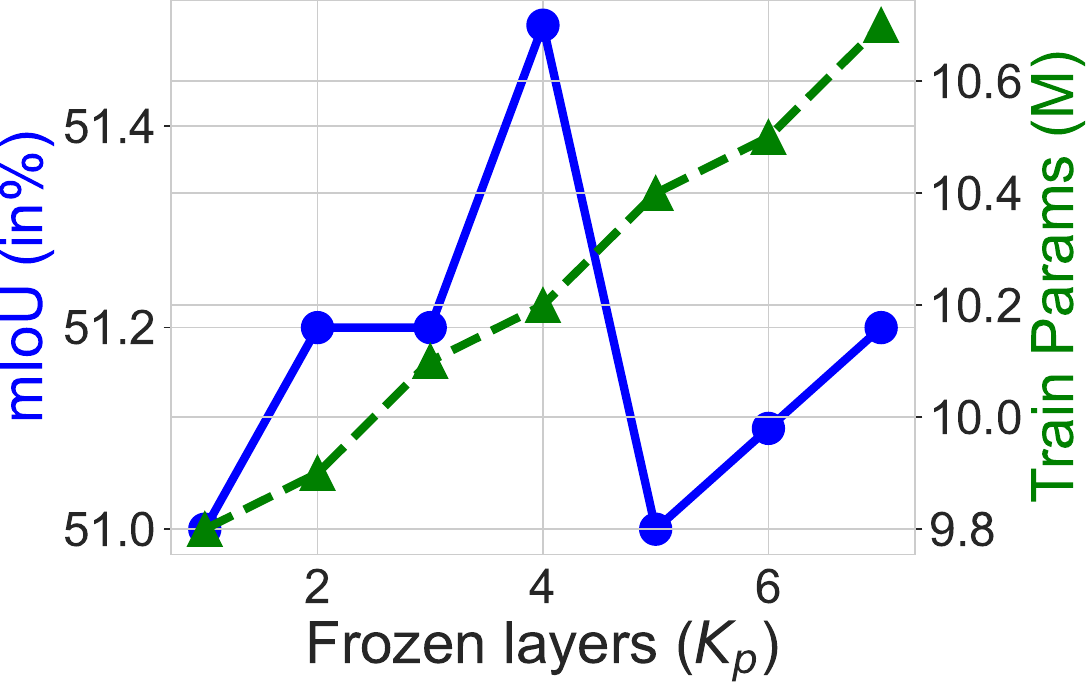}
        \caption{Frozen layers $K_p$}\label{fig:Kp}
    \end{subfigure}
    \caption{Parameter sensitivity analysis of $K_t$ and $K_p$ in  ViT-Split. The experiments are conducted using ViT-Split-S on ADE20K.} 
    \label{fig:param_sensitivity}
\end{figure}

We provide the analysis of two important hyper-parameters $K_t$ and $K_p$ in our ViT-Split, which is given in \cref{fig:param_sensitivity}. 

\textbf{Influence of $K_t$}. As shown in \cref{fig:Kt}, the mIoU initially improves when tuning between one and three layers. This improvement is likely due to the task head previously underfitting the task. However, as more layers are tuned, overall performance begins to decline, suggesting that the task head starts to overfit. This experiment demonstrates that tuning additional layers does not necessarily guarantee better performance and can easily lead to overfitting. Therefore, we opt to tune  three layers in this case.

\textbf{Influence of $K_p$}. As shown in \cref{fig:Kp}, the mIoU peaks when selecting four prior layer features. Selecting too few layers may result in missing critical information, while selecting too many can introduce noise. Additionally, we observe that increasing the number of selected layers does not increase more training parameters, highlighting the efficiency of the prior head. As a result, we choose four prior features in this case.

\subsection{Visualization}
\subsubsection{CKA analysis of other VFMs}
\begin{figure}
    \centering
    \includegraphics[width=1\linewidth]{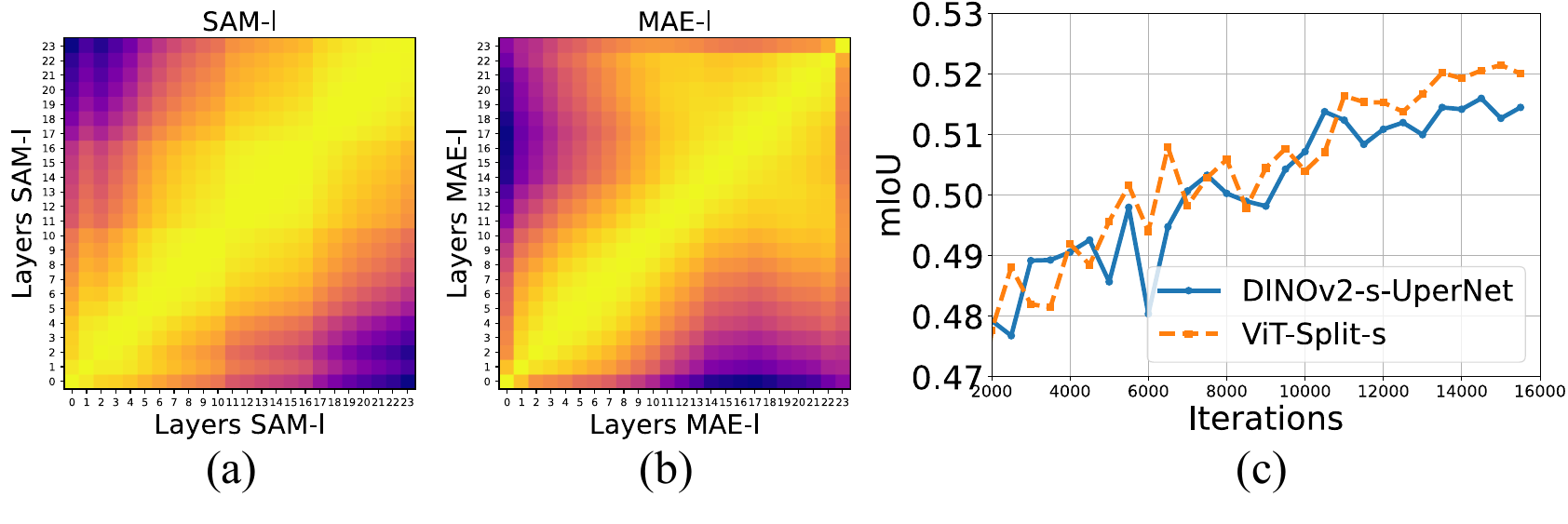}
    \caption{The CKA of SAM (a) and MAE (b). (c) Training comparison between ViT-Split-s and DINOv2-s-UperNet on ADE20K.}
    \label{fig:cka_uperbound}
\end{figure}

We also present the CKA  results for MAE-L \cite{he2022masked} and SAM-L \cite{kirillov2023segment} in \cref{fig:cka_uperbound}. The feature representations in the early layers of these VFMs exhibit similar patterns, as do those in the later layers. Based on these findings as well as those in the main paper, we hypothesize that our observation--that the layers of several VFMs can be divided into two components--may hold true for self-supervised models pretrained on large-scale dataset  (e.g., DINOv2 \cite{oquabdinov2}, MAE \cite{he2022masked}, EVA2 \cite{fang2023eva}, \textit{etc.}), as well as weakly supervised ones (say CLIP \cite{radford2021learning}, SigLip \cite{zhai2023sigmoid}, SAM \cite{kirillov2023segment}, \textit{etc.}).

\begin{figure*}
    \centering
    \includegraphics[width=1\linewidth]{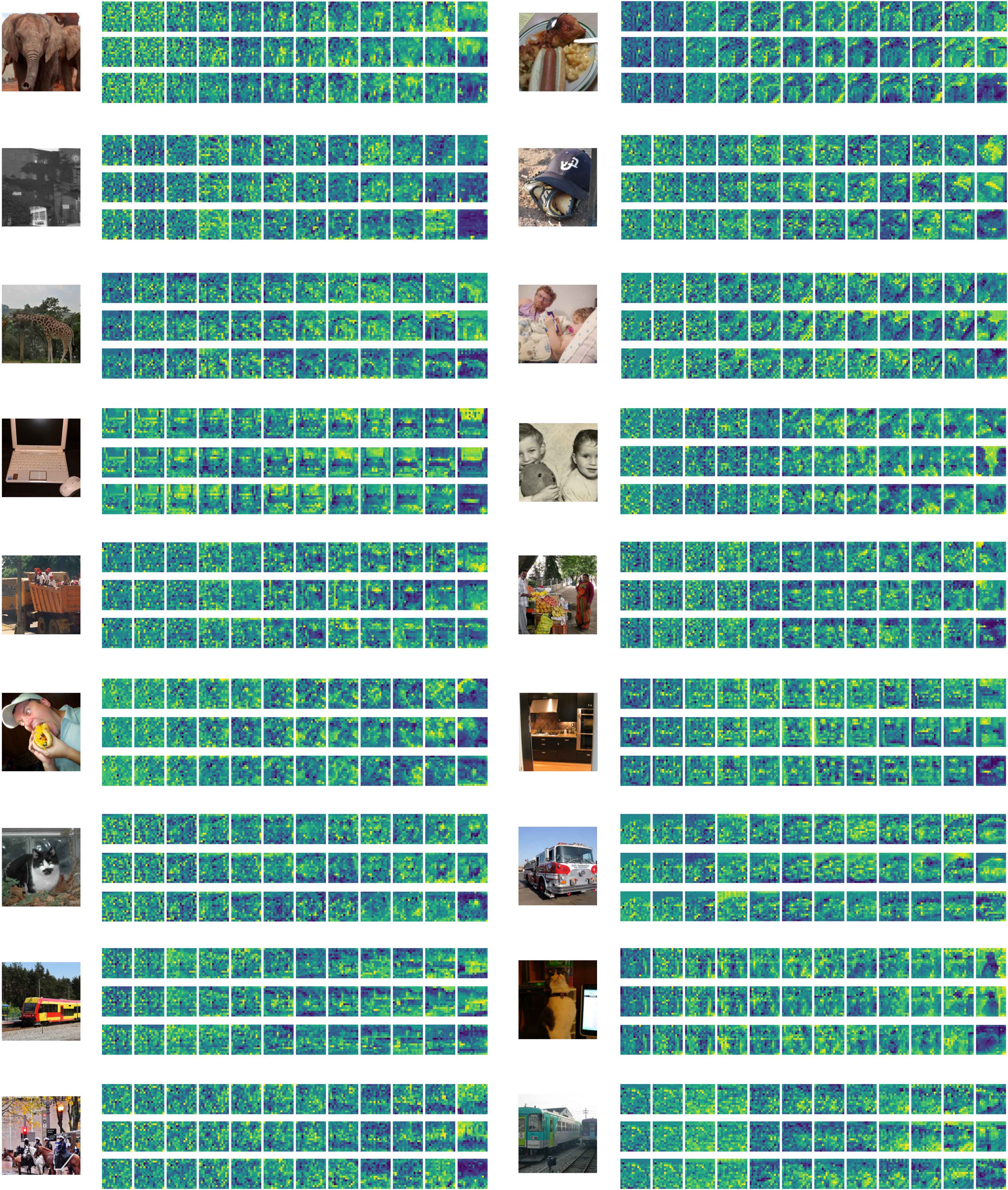}
    \caption{Further comparison of DINOv2-S layer features across original features, segmentation, and detection tasks. In each figure, the \textbf{first, second, and third rows correspond to original, segmentation, and detection features}, respectively. It can be observed that features from earlier layers exhibit similar patterns across different tasks, reflecting common low-level local features. However, features from deeper layers diverge significantly according to their specific downstream tasks.}
    \label{fig:more_layer_features}
\end{figure*}

\begin{figure*}
    \centering
    \includegraphics[width=0.83\linewidth]{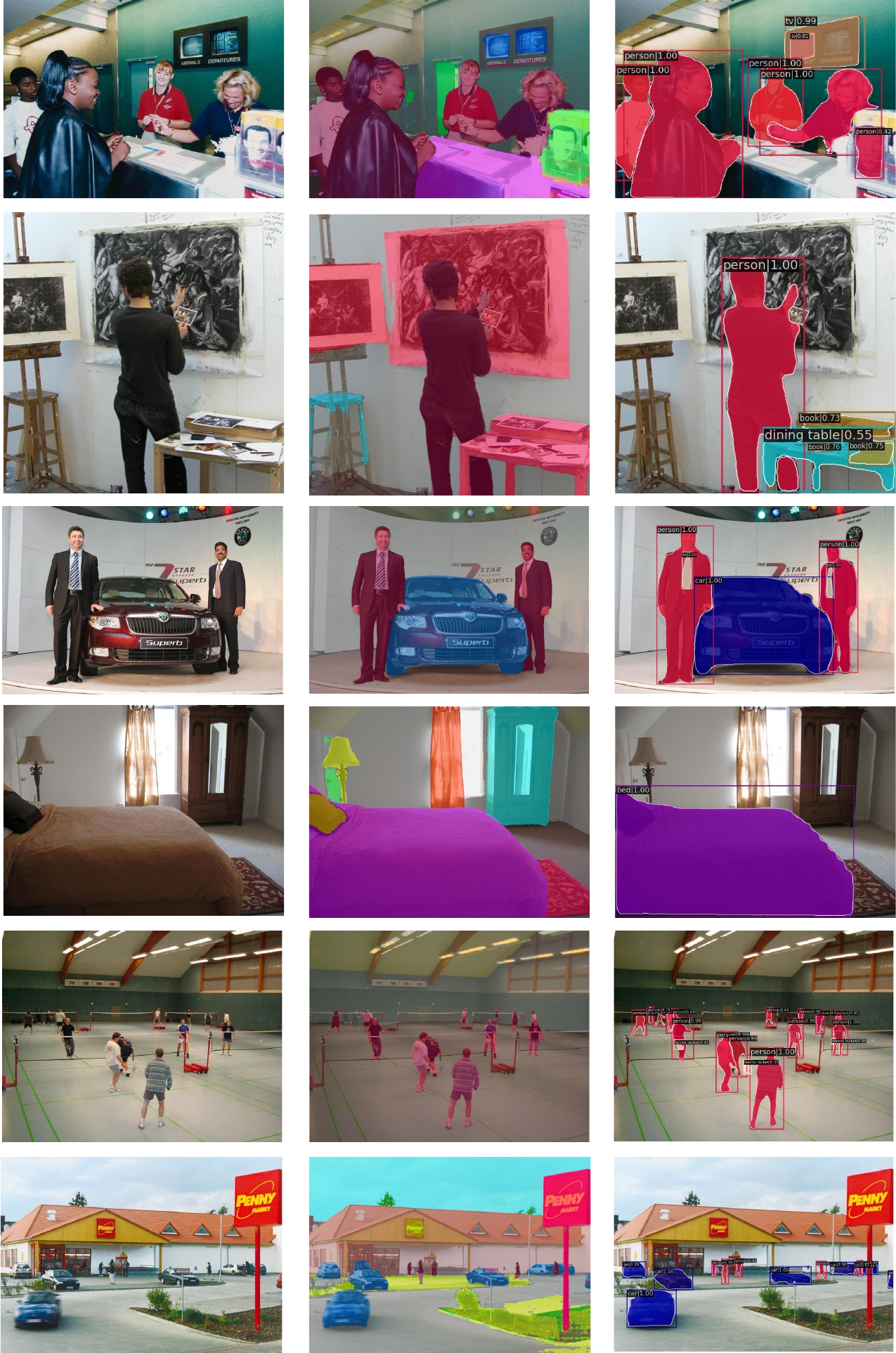}
\end{figure*}
\begin{figure*}
    \centering
    \includegraphics[width=0.80\linewidth]{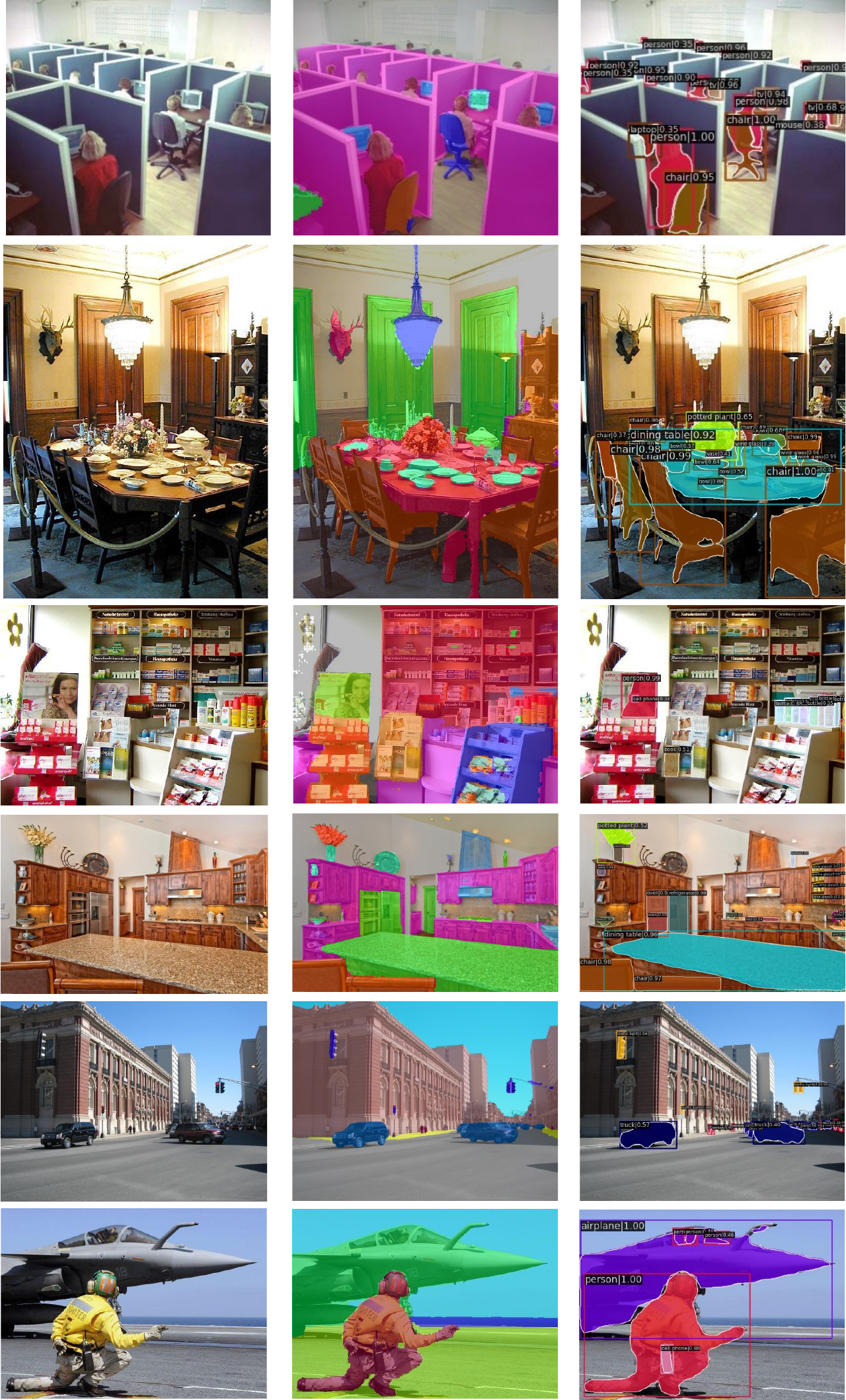}
\end{figure*}
\begin{figure*}
    \centering
    \includegraphics[width=0.83\linewidth]{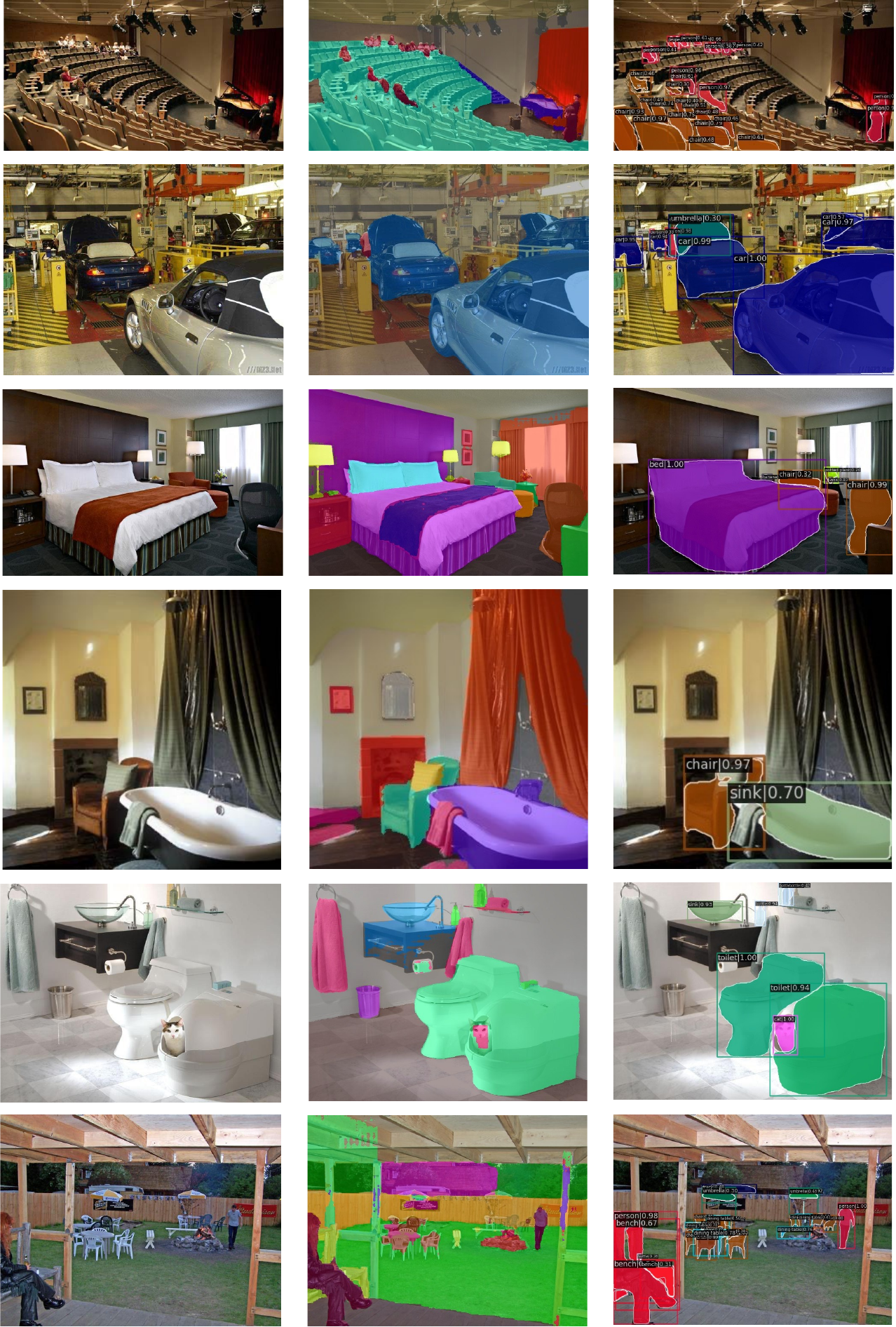}
    \caption{Semantic segmentation and instance segmentation results based on our ViT-Split-L (\textbf{left: original image, middle: semantic segmentation results, right: instance segmentation results}). }
    \label{fig:seg_det_vis}
\end{figure*}

\subsubsection{CKA analysis of different DINOv2 sizes}
\begin{figure}[h]
    \centering
    \includegraphics[width=1\linewidth]{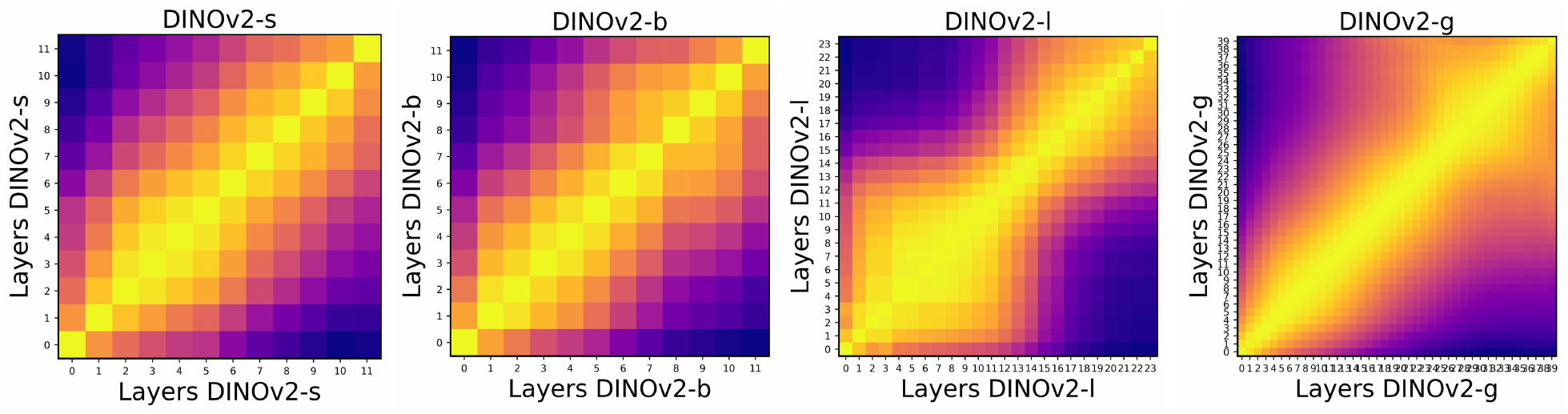}
    \caption{The CKA visualizations of different sizes of DINOv2.}
    \label{fig:dinov2_CKA}
\end{figure}

We also provide the CKA visualizations of different DINOv2 sizes in \cref{fig:dinov2_CKA}. From these visualizations, we observe that features in the early layers are more similar across different DINOv2 sizes compared to those in the later layers. As earlier mentioned, the early layers serve as an encoder to capture low-level features, while the later layers act as a decoder to produce task-specific features.

\subsubsection{More layer feature comparison}
We present additional visualizations of DINOv2 layer features across different tasks (\textit{i.e.}, DINOv2 pretraining, segmentation, and detection) in \cref{fig:more_layer_features}. These results demonstrate that earlier-layer features from various tasks consistently focus on detailed, low-level information. However, deeper-layer features diverge significantly between tasks. Specifically, features from both the original DINOv2 pretraining and semantic segmentation emphasize semantic-level information of particular objects, whereas detection features tend to highlight object corners and boundaries.

\subsubsection{Semantic segmentation and instance segmentation results}
We present semantic segmentation and instance segmentation results based on our ViT-Split-L (DINOv2 pretrained) in \cref{fig:seg_det_vis}. We utilize ADE20K and COCO2017 datasets for training these two tasks, respectively, and evaluate both on the ADE20K validation dataset.

It is worth noting that both results are obtained using the same frozen DINOv2-L backbone, meaning only the task-specific adapters and heads require training. Consequently, the overall computational cost and the number of parameters are significantly reduced compared to previous VFM-adapters, while achieving competitive or superior performance. These visualizations demonstrate the strong generalization capability of ViT-Split, highlighting its versatility, effectiveness, and efficiency across multiple downstream tasks.

\subsection{Training efficiency comparison}
\begin{table}[htbp]

    \scalebox{0.85}{    
    \begin{tabular}{c>{\columncolor[HTML]{E2F0D9}}ccc}
    \toprule
     \textbf{Type} & ViT-Split-linear & DINOv2-linear & DINOv2-UperNet\\
    \midrule
     Small & 9m25s & 15m28s & 31m21s \\
    \midrule
     Base & 17m41s & 25m16s & 40m23s \\
    \midrule
     Large & 32m49s & 44m22s & 1h19m25s \\
    \bottomrule
    \end{tabular}
    }
    \centering
    \caption{Training time comparison on ADE20K (tuning 10k iterations on 4*A6000Ada). DINOv2-linear and DINOv2-UperNet are finetuned end to end.}\label{training_time_heads}
\end{table}
To further illustrate the training efficiency compared with different heads on segmentation task, we provide the training time comparison  in \cref{training_time_heads}. For fair comparison, all of these baselines (except for ViT-Split) are finetuned using the DINOv2 backbone with two different heads (linear and UperNet) for 10k iterations on 4*A6000Ada.

From \cref{training_time_heads}, we observe that our ViT-Split reduces the training time on average of DINOv2-linear by approximately 42\% on average while maintaining the same linear head. This improvement in training efficiency is attributed to the task-head design, which \emph{prevents gradients from propagating to the early layers of the backbone}. Compared to finetuning a VFM with a larger segmentation head (DINOv2-UperNet), our ViT-Split is 2.5 times faster across three sizes on average. This highlights the huge computation overhead introduced by a large segmentation head and demonstrates the efficiency of our ViT-Split.

\subsection{Longer training time}
We try to increase the training time to illustrate the upper bound of ViT-Split.
 We conduct an experiment in \cref{fig:cka_uperbound} to explore the performance upper bound  with extended training (\textit{i.e.}, 160K iterations). As shown in \cref{fig:cka_uperbound} (c), ViT-Split-s achieves 52.2\%, improving from 51.5\% at 40K iterations and surpassing DINOv2s-UperNet (51.6\%) while maintaining faster training speeds. This demonstrates that ViT-Split can achieve better performance when training for longer time.
 
\subsection{Monocular depth estimation}

\textbf{Settings}. 
To further investigate the effectiveness of our ViT-Split, we also provide the results on monocular depth estimation (MDE) on NYU-V2 \cite{silberman2012indoor} benchmark in \cref{tab:mde_NYU}. Following \cite{li2024binsformer}, we utilize the AdamW optimizer with an initial learning rate of 3e-4 and a weight decay of 1e-2. We multiply 0.1 by the learning rate of the task head during training. Moreover, one cycle learning rate decay schedule is utilized for better performance. We train ViT-Split for 384K iterations with a total batch size of 16 on 4*A6000ada GPUs.

As shown in \cref{tab:mde_NYU}, our ViT-Split achieves competitive or even superior results compared to previous state-of-the-art methods, while using a minimal number of trainable parameters. Notably, ViT-Split employs only a single linear head rather than a specially designed head, highlighting the potential of our approach. Leveraging the prior knowledge embedded in vision foundation models (VFMs), we believe the size of the downstream task head (e.g., for depth prediction) can be further reduced to improve efficiency. 

When compared to DINOv2-G with DPT \cite{ranftl2021vision}, which uses the same DINOv2 initialization but a larger and more sophisticated head, our smaller ViT-Split-B version achieves similar performance with fewer parameters, demonstrating both the effectiveness and efficiency of our method. Furthermore, compared to traditional end-to-end fine-tuning approaches, ViT-Split achieves better performance by fully utilizing the prior knowledge inherent in VFMs. This also highlights the significant potential of large-scale self-supervised learning initialization over traditional supervised learning initialization.

\subsection{Segmentation on Pascal Context}
\textbf{Settings}. Apart from ADE20K and Cityscapes, we also provide the results on Pascal Context \cite{mottaghi2014role} in \cref{tab:seg_pascal_context}. We utilize the AdamW optimizer with an initial learning rate of 1e-4 and weight decay of 1e-2. We multiply by 0.1 to the task head during training. We train our model for 20K iterations, and the total batch size is set to 16.

As shown in \cref{tab:seg_pascal_context}, our method outperforms ViT-Adapter, achieving a 2\% improvement for the base model and a 0.3\% improvement for the large model, using just a simple linear head and training for only 20K iterations. The results demonstrate the strength of VFMs, with our method achieving both effectiveness and efficiency by fully utilizing the prior knowledge within the VFMs.
\begin{table}[t]
\centering 
\small
\setlength{\tabcolsep}{3pt} 
\scalebox{0.86}{
    \begin{tabular}{l | ccccc}
    \hline
    {Method} &  Head    & \#Train Param     & mIoU (SS/MS)     & Schedule     \\
    \hline
    ViT-Adapter-B\textsuperscript{$\dagger$} & Mask2former & 120M & 64.0/64.4 & 40k \\
    ViT-Adapter-L\textsuperscript{$\dagger$} & UperNet & 451M & 67.0/67.5 & 80k \\
    ViT-Adapter-L\textsuperscript{$\dagger$} & Mask2former & 568M & 67.8/68.2 & 80k \\
    \rowcolor[HTML]{E2F0D9}
    ViT-Split-B\textsuperscript{$\ddagger$} & Linear & 47.5M & 66.4/66.8 & 20k \\
    \rowcolor[HTML]{E2F0D9}
    ViT-Split-L\textsuperscript{$\ddagger$} & Linear & 115.8M & \textbf{68.1}/\textbf{68.6} & 20k \\
    \hline
    \end{tabular}
}
\caption{\textbf{Semantic segmentation results on the Pascal Context val with 480*480 resolution image.}  ``$\dagger$'' indicates the BEiT initialization and ``$\dagger$'' represents the use of DINOv2.}
\label{tab:seg_pascal_context}
\end{table}

\section{Limitations}
Currently, we have demonstrated the effectiveness of ViT-Split only on a limited set of VFMs, such as DINOv2 and CLIP, leaving its performance on a broader range of models to be explored in future work.

\clearpage
{
    \small
    \bibliographystyle{ieeenat_fullname}
    \bibliography{main}
}

\end{document}